\documentclass[aoas,preprint]{imsart}

\RequirePackage[OT1]{fontenc}
\RequirePackage{amsthm,amsmath}
\RequirePackage{natbib}
\RequirePackage[colorlinks,citecolor=blue,urlcolor=blue]{hyperref}

\startlocaldefs
\numberwithin{equation}{section}
\theoremstyle{plain}

\endlocaldefs
\usepackage{graphicx}
\usepackage{amsfonts}
\usepackage{amsmath}
\usepackage{graphicx}
\usepackage{nicefrac}
\usepackage[a4paper]{geometry}
\usepackage{nicefrac}
\usepackage[linesnumbered,ruled,vlined]{algorithm2e}
\usepackage{multirow}
\begin{document}

\begin{frontmatter}
% \title{Quantifying variability in estimated functional connectivity networks
% via mixed neighbourhood selection}
% \title{Inferring brain connectivity networks from functional MRI data via mixed neighbourhood selection}
% \title{Decoding population and subject-specific contributions to brain
% connectivity networks via Mixed Neighborhood Selection}
% \title{Learning population and individual contributions to brain connectivity 
% via Mixed Neighborhood Selection}
\title{learning population and subject-specific brain connectivity networks via mixed neighborhood selection}
\runtitle{Mixed neighborhood selection}

\begin{aug}
\author{\fnms{Ricardo Pio} \snm{Monti}\thanksref{m1}\ead[label=e1]{ricardo.monti08@imperial.ac.uk}},
\author{\fnms{Christoforos} \snm{Anagnostopoulos}\thanksref{m1}\ead[label=e2]{canagnos@imperial.ac.uk}}
\and
\author{\fnms{Giovanni} \snm{Montana}\thanksref{m1,m2}
\ead[label=e3]{giovanni.montana@kcl.ac.uk}
\ead[label=u1,url]{http://www.foo.com}}

% \thankstext{t1}{Some comment}
% \thankstext{t2}{First supporter of the project}
% \thankstext{t3}{Second supporter of the project}
\runauthor{R. P. Monti et al.}

\affiliation{Imperial College London\thanksmark{m1} and King's College London\thanksmark{m2}}

\address{R. P.  Monti, C. Anagnostopoulos\\Department of Mathematics, \\Imperial College London\\
SW7 2AZ, London, UK\\
\printead{e1}\\
 \phantom{E-mail:\ }\printead*{e2}}
 
\address{G. Montana\\Department of Biomedical Engineering,\\ King's College London,\\ St Thomas' Hospital, London SE1 7EH, UK\\
\printead{e3}\\
%\printead{u1}
}
\end{aug}

\begin{abstract}
% In neuroimaging data analysis, Gaussian graphical models 
% are 
% often used to model
% statistical dependencies across 
% spatially remote brain regions known as functional connectivity.

% Understanding latent structure across multiple, related 
% Gaussian Graphical models (GGMs) is a challenging statistical problem
% with important applications in neuroimaging. 
% In the latter, GGMs are often used to model
% statistical dependencies across spatially remote brain regions; referred to as 
% functional connectivity. 
In neuroimaging data analysis, Gaussian graphical models 
are 
often used to model
statistical dependencies across 
spatially remote brain regions known as functional connectivity.
Typically, data is collected across a cohort of subjects
and the scientific objectives consist of 
estimating 
population and subject-specific graphical models. 
A third objective that is often overlooked
involves quantifying inter-subject variability
and thus 
identifying regions or sub-networks 
that demonstrate heterogeneity across subjects.
Such 
information is fundamental in order to thoroughly understand 
the human connectome.
We propose \textit{Mixed Neighborhood Selection}
in order to simultaneously address the three
aforementioned objectives.
By recasting covariance selection as a neighborhood selection 
problem we are able to efficiently learn
the topology of each node. 
We introduce an additional mixed effect component to 
neighborhood selection in order to 
simultaneously estimate a graphical model for the population 
of subjects as well as for each individual subject.
% The use of mixed effect models allows us to untangle the 
% the population and subject-specific contributions.
% Furthermore,
% such an approach also allows us to 
% % we are also able to 
% highlight edges which are highly variable within a population.
% We extend traditional neighbourhood selection 
% by incorporating a random effect component, thereby directly modeling 
% inter-subject heterogeneity. 
The proposed method is 
validated empirically through a series of simulations and applied to 
resting state data for healthy subjects taken from the ABIDE consortium.
\end{abstract}


% \begin{keyword}
% \kwd{sample}
% \kwd{\LaTeXe}
% \end{keyword}

\end{frontmatter}

\section{Introduction}
\label{sec--Intro}

At the forefront of neuroscientific research is the study of
functional connectivity; defined as the statistical dependencies across
spatially remote brain regions \citep{friston1994functional, friston2011functional}. 
While traditional neuroimaging studies focused on the 
roles of specific brain regions, there has recently 
been a significant shift towards  understanding the %the study of 
connectivity across regions \citep{smith2012future}.
This shift has been partially catalyzed by recent advances in 
imaging techniques. 
In particular, the introduction of 
functional MRI (fMRI) has 
played a crucial role
%greatly boosted the study of
%functional connectivity 
%as it provides
by providing a non-invasive mechanism through
which to obtain whole-brain coverage of neuronal activity \citep{huettel2004functional, poldrack2011handbook}. 
The focus of this work involves estimating functional connectivity networks from fMRI data, 
however the methodology presented can also be used in conjunction with other imaging modalities. 

From a statistical perspective, Gaussian Graphical models (GGMs) are 
often employed to model functional connectivity \citep{smith2011network, varoquaux2013learning}.
In this manner, undirected connectivity networks can be inferred by studying 
the conditional independence structures across brain regions \citep{lauritzen1996graphical}. 

However, there are several caveats of neuroimaging data which must be adequately considered in order to
accurately estimate functional connectivity networks, principal among which is 
the \textit{replicated} nature of imaging experiments. 
% Here we allude to the fact that 
Neuroimaging datasets typically consist of multivariate time series data collected
across a cohort of subjects. 
As a consequence,
the goal is often to both learn 
multiple \textit{replicated} brain connectivity networks across a cohort of subjects as well 
as understand the properties and characteristics which define the population. 
Hereafter we refer to such data as \textit{replicated} data.
% and note that
% it is particularly relevant for the latter objective.

% Neuroimaging data typically
% consists of multiple replicates across 
% possibly many subjects.
% % . As a result, neuroimaging datasets 
% % consist of many related observations. 
% Hereafter, we refer to such data as \textit{replicated} data.
% in contrast to 
% \textit{individual} data for which GGMs are often 
% which is often the focus of statistical 
% research, especially in the context of GGMs. 

% The predominant focus of this work is dealing with one of these aspects that is 
% often overlooked within the neuroimaging community. 

% \newpage

% Understanding latent structure across multiple, related Gaussian Graphical models (GGMs) is a 
% challenging statistical problem
% with important applications in modern neuroimaging. 
% In the latter, GGMs are often used to 
% model functional connectivity, defined as 
% the statistical dependencies present across
% spatially remote brain regions \citep{smith2013functional}. 
% A unique aspect of neuroimaging data is that it typically
% consists of multiple replicates across 
% possibly many subjects. As a result, neuroimaging datasets 
% consist of many related observations. 
% We refer to such data as \textit{replicated} data in contrast to 
% \textit{individual} data which is often the focus of statistical 
% research. 

Two further hallmarks of neuroimaging data
are its high dimensional nature as well as its reproducibility.
Particularly in the context of connectivity networks, observed patterns across large numbers of 
brain regions have been shown to be consistent 
across subjects and scanning sessions
\citep{damoiseaux2006consistent, zuo2010oscillating}. 
In recent years, these properties have 
motivated novel methodological advances which seek to more 
reliably estimate subject-specific functional connectivity networks by leveraging information 
across a population \citep{ varoquaux2010brain, wee2014group}.
However, 
from a
neuroscientific perspective several problems remain; chief of which is 
understanding and quantifying
inter-subject variability \citep{kelly2012characterizing, mueller2013individual}. 

Within the neuroimaging literature, standard approaches to studying \textit{replicated} data 
can be summarized in two main avenues of research. The first involves learning a 
separate GGM for each subject independently. While such an approach allows
researches to examine subject-specific hypotheses, 
 it may also prove problematic in the context
of  neuroimaging data. This is  particularly true when studying 
fMRI %functional MRI (fMRI) 
data, which is characterized
by its high spatial and low temporal resolution. 
Much of the neuroimaging literature looks to address this issue via the introduction 
of regularization constraints \citep{smith2011network, ryali2012estimation, lee2011sparse, wehbe2014regularized},
however such methods may still perform poorly if only a limited  number of observations are available per subject.
% Further, functional connectivity networks demonstrate reproducible properties across subjects thereby
% suggesting that more reliable network estimates may be obtained by 
% leveraging information across subjects. 
Notwithstanding, a practical advantage such a strategy is that variability across a 
population can be quantified \citep{varoquaux2013learning}. However, 
the resulting two-step procedure is ultimately limited by the fidelity of the 
estimated individual networks. 
% the resulting 
% two-step procedure is difficult to justify. 

The second approach is to learn a single GGM
that is representative of the entire population of brain networks. 
In addition to greatly facilitating the interpretation of results, this 
strategy leverages information across subjects (albeit in a naive manner), thereby
alleviating issues caused by the high dimensional nature of the data. 
The most significant deficiency
stems from the fact that it fails to adequately 
model the inter-subject variability present in neuroimaging data \citep{poldrack2011handbook, ashby2011statistical, mueller2013individual}. 
Furthermore, the question of understanding variability across the population is often sidelined
thereby undermining the interpretability of results \citep{fallani2014graph}. 

The objective of this work is to reconcile the two popular approaches presented above; thus
allowing for accurate network estimation at subject-specific and population levels while also 
quantifying variability present across a cohort. 
In recent years significant progress has been
made towards these objectives,
in particular through the use of novel regularization schemes. 
For example, the \textit{graphical lasso} penalty \citep{friedman2008sparse} 
has been used extensively to 
recover the sparse support on a subject-specific basis \citep{varoquaux2013learning}.
An alternative approach involves sharing information
across subjects via employing either \textit{group} or \textit{fused}
lasso penalties \citep{varoquaux2010brain, danaher2014joint}. % wee2014group}. 
Based upon the notion that 
resting state networks demonstrate reproducible properties across
subjects, such approaches look
to alleviate the high-dimensional nature of fMRI data by 
effectively leveraging data. 
This recent work has clearly
demonstrated the benefits obtained 
through leveraging information across a cohort of subjects.
However, while such approaches constitute 
significant improvements both from a methodological perspective, the question of quantifying
inter-subject variability remains 
unanswered.

In this work we present a novel methodology through which to partially 
address these issues. The proposed method, named
Mixed Neighborhood Selection (MNS), is based on 
the neighborhood selection method introduced by \cite{meinshausen2006high}.
By recasting covariance selection as a series of linear
regression problems, neighborhood selection methods are able to
learn the local network topology of each region. {MNS} extends
neighborhood selection by incorporating an additional
random effect component. 
This component is introduced with the intention of 
learning both a population topology (captured in the fixed effect terms)
as well as subject-specific topology (captured in the random effects) for each node.
This serves to directly model 
inter-subject variability and provides a much richer model of functional connectivity. 
In particular, 
the proposed method is able to
partition edges according to their reproducibility across the cohort.
% into three groups; present, absent or variable. 
In doing so, 
{MNS} provides an additional layer of information which can be exploited 
to further understand functional connectivity.
Moreover, by effectively differentiating between 
reproducible edges present across the entire cohort and 
highly variable edges, the proposed method is able to
leverage information in a discriminative manner,  leading to more
reliable network estimates.

\begin{figure}[h!]
\begin{center}
\includegraphics[width=.8\textwidth]{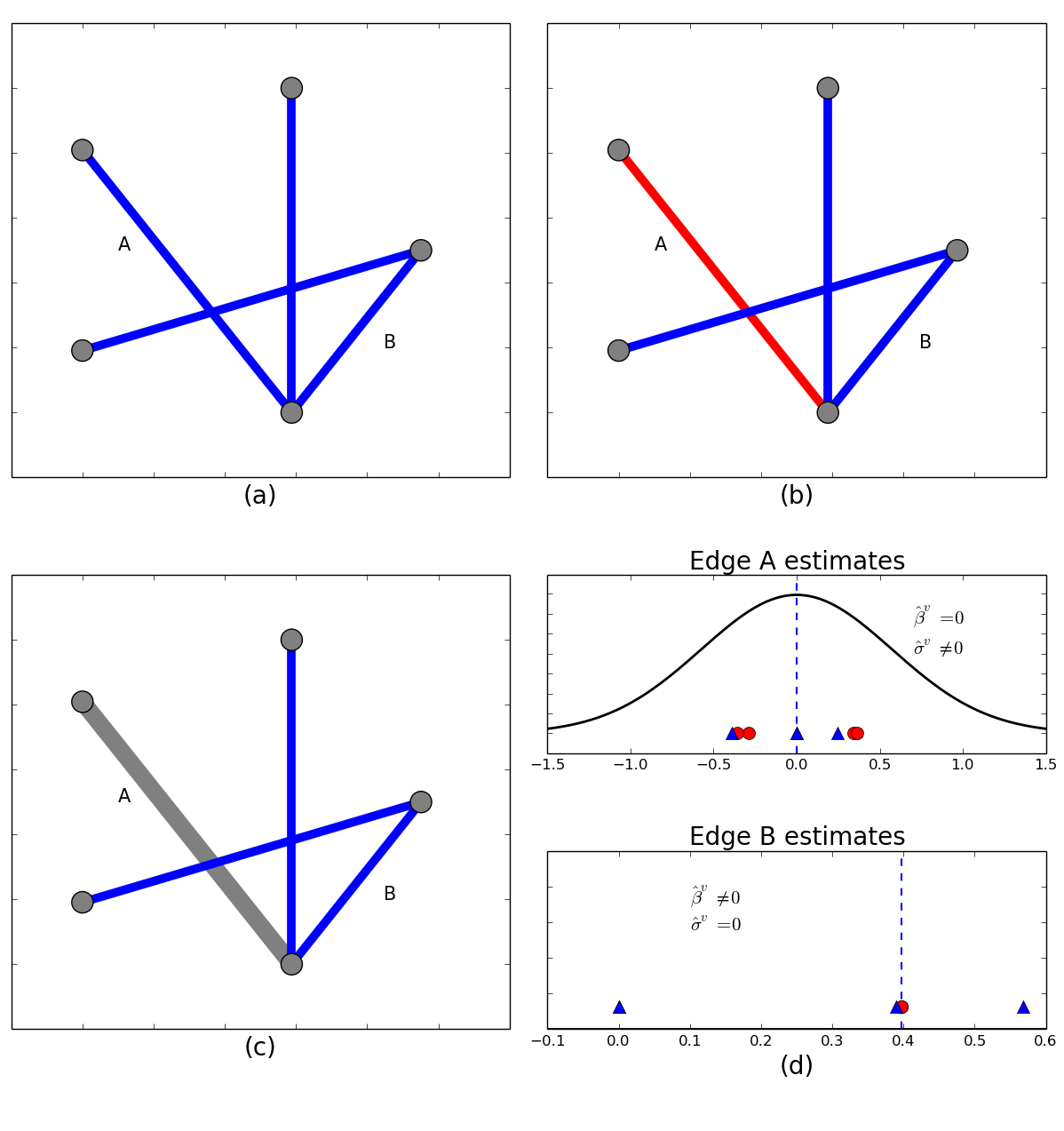}
\caption{Toy motivating example to illustrate the capabilities of the proposed method. Networks were simulated with $p=5$ nodes
and with $n=8$ observations per subject for $N=4$ subjects. The functional connectivity 
networks for two of the subjects is shown in \textbf{(a)} while the networks for the remaining two 
is shown in \textbf{(b)}.
Blue edges indicating positive partial correlations
and red edges indicative of negative partial correlations. 
% We note that there is a significant proportion of edges shared
% across all subjects, but there is one edge in particular which is variable. 
A significant proportion of the edges are shared across subjects with a single variable edge.
The results for our proposed
method are shown in \textbf{(c)}; blue lines indicate edges shared by the entire population while thick gray edges 
indicate highly variable edges. Estimated edge coefficients for edges A and B are shown as obtained 
by the {MNS} algorithm as well as by applying the graphical lasso to each dataset independently in \textbf{(d)}:
Dashed blue lines 
indicate the estimated population edge value while the 
% are the population edge value while 
solid back line is the estimated probability density function of that edge based on the random effects (i.e.,
it visualizes the BLUP distribution). 
Blue, triangular points indicate edge values as estimated by the graphical lasso while red, circular points
indicate subject-specific {MNS} estimates (i.e., the BLUPs).}
\label{motExample}
\end{center}
\end{figure}

In order to illustrate the capabilities of the proposed method we present a brief
motivating example, shown in Figure [\ref{motExample}]. We consider a scenario where there is a population consisting of
four individuals whose
functional connectivity networks share a common structure but also demonstrate some variability. In 
particular, one edge varies across subjects such that two subjects exhibit the functional connectivity
shown in Figure [\ref{motExample}a] and the remaining two Figure [\ref{motExample}b]; the edge in question (edge A) is 
shown to vary from positive to negative across groups. In such a scenario,
it is of scientific interest both to uncover the correct functional connectivity networks as well as to
correctly identify edges which are variable within the population. This is precisely what MNS
is capable to achieving.
The results are shown in Figure [\ref{motExample}c] where the blue lines indicate edges shared across
the entire population.
The thick  gray edges indicate \textit{random effect} edges that demonstrate high variability. 
Figure [\ref{motExample}d] shows the estimated edge coefficients for 
two edges of interest when estimated using the proposed method and 
the graphical lasso.
We note that in addition to correctly recovering the 
sparsity structure, the proposed method is able to 
discriminate edges according to their reproducibility across the cohort.
% report edges which are reproducible across the entire cohort. 
% report which edges are 
% variable and which are reproducible across
% the entire cohort. 
This is in contrast
to what could be achieved by studying 
the networks estimated for each
subject independently. This point is demonstrated in Figure [\ref{motExample}d] where
the estimated edge coefficients for the graphical lasso\footnote{
the graphical lasso was run independently for
each subject. The regularization parameter for each subject was selected using cross-validation} are shown to be
highly variable across both edges, one of which is does not vary across subjects. As a result, 
it follows that identifying variable edges in a two-step procedure is challenging,  even in low-dimensions.

The remainder of this manuscript is organized as follows. 
Prior to discussing the proposed method, we briefly  
review current methodology in neuroimaging in Section \ref{sec--Overview}.
The proposed is then detailed in Section \ref{sec--Methods}. 
% In Section \ref{sec--Methods} 
% we briefly review current approaches within the neuroimaging community before introducing 
% the proposed method. 
In Section \ref{sec--SimStudy}
we present an extensive simulation study and the proposed method is applied to resting-state fMRI data 
from the ABIDE consortium in Section \ref{sec--APP}. 

\section{Current methodology in Neuroimaging}
\label{sec--Overview}

In this section we outline current methodology within the neuroimaging community. 
To set notation, we assume
% We assume 
we have access to fMRI time series across a cohort of $N$ subjects. 
For the $i$th subject, it is assumed we observe an $n$-dimensional 
fMRI time series across $p$ 
fixed regions of interest. 
We write $V=\{1, \ldots, p\}$ to denote the
set of regions or nodes and refer to the 
dataset 
% We therefore denote the dataset 
for the $i$th subject by $X^{(i)} \in \mathbb{R}^{n \times p}$.
Further, we write $X^{(i)}_v \in \mathbb{R}^{n \times 1}$ to denote the 
time-series for any node $v \in V$. Similarly, we let $X^{(i)}_{\backslash v} \in \mathbb{R}^{n \times (p-1)}$ denote 
the times-series across all remaining nodes.

Throughout this work it is assumed that the data of each subject 
follows a stationary multivariate Gaussian distribution. Since our primary interest is 
the estimation of functional connectivity networks, summarized in the inverse covariance matrix, we 
assume without loss of generality that each $X^{(i)}$ has zero mean. As a result, we have that 
$X^{(i)} \sim \mathcal{N}(0, \Sigma^{(i)})$ 
where $\Sigma^{(i)}$ is the covariance for the $i$th subject.

% Prior to introducing the proposed method, we briefly review current approaches within the 
% neuroimaging community  and discuss challenges associated with the study of fMRI data
% in Section \ref{sec--Review}. 
% The proposed method is then detailed in Section \ref{sec--Proposed}. Finally, related work is discussed in 
% Section \ref{sec--RelWork}.

\subsection{Modeling connectivity through GGMs} %Learning graphical models}%Current methodology and challenges in neuroimaging}
\label{sec--Review}
Under the assumption of Gaussianity, estimating functional connectivity networks is 
equivalent to learning the conditional independence structure for each subject.
This can be succinctly represented as a graphical model, $G^{(i)}=\left (V,E^{(i)} \right )$, where
the edge set, $E^{(i)}$, encodes conditional dependencies across a fixed set of nodes, $V$
\citep{lauritzen1996graphical}.
Formally, the edge set summarizes the non-zero entries in the precision matrix, thus:
\begin{equation}
E^{(i)} = supp \left ( \left (\Sigma^{(i)} \right )^{-1} \right ) = \left \{ (j,k) : \left ( \Sigma^{(i)} \right )^{-1}_{j,k} \neq 0 \right \}. 
\label{SubjectEdgeSupport}
\end{equation}
The resulting edges are taken to be indicative of functional relationships between 
spatially remote regions of the brain, allowing us to interpret the 
estimated graphical model as a functional connectivity network. 

The problem of learning the aforementioned dependence structures, first 
posed by \cite{dempster1972covariance}, is typically referred to as
covariance selection. 
This is a challenging problem that is further exasperated by the high-dimensional
nature of fMRI data. As a result, regularization is often introduced
\citep{smith2011network, varoquaux2013learning}. 
% While initially covariance selection problems where solved 
% using greedy algorithms, several highly efficient methods based on 
% optimizing convex functions have since been proposed \citep{ friedman2008sparse}.
In particular, the use of neighborhood selection methods introduced by \cite{meinshausen2006high}
have been widely adopted within the neuroimaging community \citep{lee2011sparse, wee2014group, pircalabelu2015focused}.
As these methods will form the backbone for the proposed method, we formally discuss neighborhood selection below. 

\subsubsection{Neighborhood selection}
\label{sec--NghSel}
The intuition behind neighborhood selection stems from the fact that we may learn the conditional
independence structure across all nodes by iteratively learning the independence structure of each node.
The latter is referred to as the neighborhood for each node $v \in V$. 
We write  $ne^{(i)}(v)$ to denote the estimated neighborhood of node $v$ at the  $i$th subject. 
% For each node, $v \in V$, the latter is referred to as 

\cite{meinshausen2006high} propose to learn the neighborhood of each node $v \in V$ by
considering the optimal prediction of $X_v^{(i)}$ given the time series of the remaining nodes. 
This results in the formulation of the following linear model for node $v$:
\begin{equation}
 X_v^{(i)}  = X_{\backslash v}^{(i)} \beta^{(i), v} + \epsilon^{(i), v}
\end{equation}
where $\epsilon^{(i), v} \sim \mathcal{N}(0, \sigma^2 I)$
is white noise. In such a regression model it follows that nodes that are not in the neighborhood of $v$ will be omitted
from the set of optimal predictors. 
Thus neighborhood selection can be reformulated as subset selection 
in a linear model.
The latter problem has received considerable attention; one notable solution being that of the Lasso \citep{tibshirani1996regression}.  
Briefly, the Lasso imposes a constraint on the  $\ell_1$ norm of the regression coefficients. 
This allows the Lasso to 
obtain parsimonious solutions %(some regression coefficients will be shrunk to zero)
whilst remaining  convex. 

The neighborhood selection approach described in \cite{meinshausen2006high} proceeds
by solving the following convex optimization problem for each node $v$:
% as follows, 
% for each node $v \in V$ solve the following convex optimization problem:
\begin{equation}
 \hat \beta^{(i), v} = \underset{\beta^{(i), v} \in \mathbb{R}^{p-1} }{\operatorname{argmin}}  \left \{ \frac{1}{2} ||X_v^{(i)} - X_{\backslash v}^{(i)} \beta^{(i),v}||_2^2 + \lambda || \beta^{(i), v}||_1\right \}.
 \label{NghSelEq}
\end{equation}
Due to the parsimony property of the Lasso, some elements of $\hat \beta^{(i), v} \in \mathbb{R}^{p-1}$ will be shrunk
to zero, effectively removing these nodes from the optimal prediction set. An estimate
for the neighborhood of $v$ is subsequently 
defined as:
\begin{equation}
 \hat{ne}^{(i)}(v) = \left \{ u \in V\backslash\{v\} : \hat \beta^{(i), v}_u \neq 0  \right \}.
\end{equation}
That is to say, the neighborhood of $v$ is the set of all nodes included in the Lasso solution to equation (\ref{NghSelEq}).
Given an estimate of the neighborhood of all nodes, the edge structure for a graphical model
can then be obtained using one of the following rules \citep{meinshausen2006high}:
\begin{equation}
\small
 E^{(i)} _{{OR}} = \{ (v,u) : u \in \hat \beta^{(i),v} \mbox{ \textbf{or }} v \in \hat \beta^{(i), u}  \} \hspace{1.5mm} \mbox{or} \hspace{1.5mm} E^{(i)}_{{AND}} = \{ (v,u) : u \in \hat \beta^{(i), v} \mbox{ \textbf{and }} v \in \hat \beta^{(i), u}  \}.
\label{ANDOR}
\end{equation}
% The use 
% of either of these rules has been shown to lead to
% consistent edge selection. In the remainder we employ the
% more conservative AND rule. 

% Neighborhood selection methods have been extensively employed within the neuroimaging community.
% Pertinent examples include the work of \cite{lee2011sparse}, who employ neighborhood selection 
% to discover differences in network structure 
% across healthy controls and patients suffering from autism spectrum disorder. A similar approach is 
% described in \cite{wee2014group}, who study the effects of mild cognitive impairment on functional connectivity. 

% Furthermore, \cite{pircalabelu2015focused} recently employed 
% neighborhood selection to perform graphical model selection in the context of fMRI data.

% propose a novel method through which to select regularization parameters
% in the context of fMRI data. 

\subsubsection{Persistent challenges}
A fundamental challenge associated with estimating functional connectivity
networks is that the accuracy of the estimated graphical models 
remains limited when the number of observations, $n$, is small relative to the dimensionality
of the data, $p$. 
This is particularly concerning in the context of fMRI data which is renown for its
high spatial and low temporal resolution \citep{poldrack2011handbook}.

While the introduction of regularization is able to partially address this concern,
such methods will still perform poorly in the presence of limited observations (see for example the results
of our motivating example
presented in Figure [\ref{motExample}]). Moreover, there is no consensus within the
neuroimaging community regarding 
the \textit{true} sparsity level of functional connectivity networks. 
Some studies suggest that connectivity networks have evolved 
to achieve high efficiency of information transfer at a low connection cost, 
leading to sparse networks \citep{bullmore2009complex}.
However, convincing arguments have also been posited against sparsity.  For example,
\cite{markov2013cortical} propose a high-density model
where efficiency is achieved via the presence of highly heterogeneous edge strengths between nodes.
As such, it is unclear how sparsity inducing regularization techniques serve the 
inferential goals of cognitive neuroscience \citep{wehbe2014regularized}. This is particularly true in the 
context of functional connectivity.

A naive solution to this concern is to concatenate observations across all subjects and estimate a
single graphical model. 
While such approaches may lead to interpretable results, unless inter-subject 
heterogeneity is adequately modeled the resulting network will be a poor representation of 
the true population covariance structure \citep{ashby2011statistical}. 
% While such approaches lead to interpretable results, inter-subject
% variability is not adequately modeled. 
% In the presence of a highly heterogeneous group of subjects, the estimated 
% network may therefore be a poor representation of the true connectivity \citep{ashby2011statistical}. 

Recently, more sophisticated methods have sought to address this issue by exploiting 
the fact that functional connectivity networks
are often reproducible across subjects \citep{varoquaux2010brain, guo2011joint, danaher2014joint}; thereby suggesting 
that improved connectivity estimates could be obtained 
via leveraging information across a cohort of subjects. 
Such methods effectively regularize the differences between estimated networks across multiple subjects via the introduction 
of an additional penalty. 
For example, by imposing a group lasso penalty \cite{varoquaux2010brain} are able to enforce a 
common support structure on the graphical models of each subject. 

However, as we look to demonstrate in the remainder of this work, 
these approaches suffer two fundamental deficiencies. 
First and foremost, they do not quantify variability across a cohort of subjects. 
This is a topic which has not received sufficient attention within 
the neuroimaging community and is fundamental for the understanding and 
interpretation  of estimated networks \citep{fallani2014graph}. 
Second, the aforementioned methods leverage information across subjects in a 
homogeneous fashion.
%non-discriminative manner.
As we will demonstrate, the proposed method is able to discriminate edges according to their 
reproducibility. This information then serves to guide the proposed method when 
leveraging information; allowing for 
less stringent regularization to be enforced upon variable edges.

\section{Methods}
\label{sec--Methods}
% \label{sec--Proposed}

% As highlighted in Section \ref{sec--Overview}, 
There are a number of challenges and 
objectives which must be considered when studying neuroimaging data across a cohort of subjects.
Predominant objectives include learning population and subject-specific 
connectivity networks. 
In this section we present the proposed method, termed Mixed Neighborhood Selection (MNS),
through which to address both these objectives together with a third that is often 
overlooked; 
that of understanding the variability in covariance structure across a population. 
% As a result, 
% the objectives of the proposed method are as follows:
% \begin{enumerate}
%  \item Accurately estimate the population covariance structure%, $E^{pop}$.
%  \item Accurately estimate the subject-specific covariance structure%. 
% %  This 
% %  corresponds to accurately learning the subject-specific idiosyncrasies, captured by
% %  $\tilde E^{(i)}$. 
%  \item Reliably quantify variability present across a cohort. In order to obtain 
%  insightful and interpretable results, we look to quantify variability 
%  on an edge-by-edge basis. %This corresponds to recovering $\tilde E$. 
% \end{enumerate}

% The goal of this work is to address the three aforementioned
We look to address these 
objectives by adequately modeling inter-subject heterogeneity. 
This is achieved via proposing a novel model for the covariance structure across a cohort 
of subjects. The proposed model, which looks to categorize edges on the basis
of their reproducibility, is described in Section \ref{sec--CovModel}.
This in turn 
allows MNS to leverage information across subjects in a discriminative manner, described in 
Section \ref{sec--MNS}. 
An efficient estimation algorithm is outlined in Section \ref{sec--EMalgo}. 
Parameter tuning is discussed  in Section \ref{sec--ParamSelect} and 
Section \ref{sec--RelWork} contains a discussion of
related approaches from the literature.

% In order to achieve this
% we being by proposing a novel model for covariance structure across a cohort of subjects, described
% in Section \ref{sec--CovModel}.
% The proposed model looks to separate edges on the basis of their reproducibility across
% subjects, thus separating stable edges from highly variable ones. This in turn 
% allows the proposed method to leverage information across subjects in a discriminative manner, described in 
% Section \ref{sec--MNS}. 
% An efficient estimation algorithm is outlined in Section \ref{sec--EMalgo}. 
% Finally, parameter selection is discussed  in Section \ref{sec--ParamSelect}.

\subsection{A novel covariance model}
\label{sec--CovModel}

We propose to model the covariance structure for each subject as the union of a shared covariance structure 
together with subject-specific idiosyncrasies. 
This corresponds to the assumption that there exists a shared covariance structure which manifests itself
across all subjects together with subject-specific deviations from this structure. The latter 
allows our model to accommodate inter-subject variability which cannot be ignored.
% Thus, we assume there is a 
% shared covariance structure which manifests itself across all subjects. 
% However, we also wish to account for inter-subject 
% variability which cannot be ignored. 
As a result, we model 
the conditional independence structure of each subject 
as the union of the support of a sparse population network 
and a subject-specific network. 
% The former accounts for covariance structure which is reproducible across
% the entire cohort of subjects while it is the latter which is able to
% explain subject-specific deviations. 
Formally, the support for each subject's conditional independence structure, originally defined in 
equation (\ref{SubjectEdgeSupport}), is modeled as:
\begin{equation}
 \label{CovEq}
%  supp \left ({\Sigma^{(i)}}^{-1} \right ) =  supp~ \Big (\Theta^{pop} \Big) \cup supp \left ( \Theta^{(i)} \right)
E^{(i)} =  E^{pop} \cup \tilde E^{(i)}%supp \left ( \Theta^{(i)} \right)
\end{equation}
Here we interpret $E^{pop}$ as the population edges which encode the conditional 
independence structure shared across the entire population.
Under the assumption of Gaussianity, it follows that $E^{pop}$ is associated 
with a population precision matrix, $\Theta^{pop} \in \mathbb{R}^{p \times p}$. 
On the other hand, it is $\tilde E^{(i)}$ which encodes subject-specific 
deviations from the population covariance structure. 
We define $\tilde E = \bigcup_{i=1}^N \tilde E^{(i)}$ as the set of edges 
demonstrating variability across the entire population. 
This variability 
may either be attributed to the nature of the edge (i.e., positive or negative partial correlations 
as in the motivating example described in Figure [\ref{motExample}]) or partial presence of the edge (i.e., the edge is only
present within a sub-group of subjects). 

The objective of the proposed method therefore corresponds to 
accurately identifying 
both $E^{pop}$ and $\tilde E^{(i)}$. % and $\tilde E$.
Given $E^{pop}$ and  $\tilde E^{(i)}$,
one  can infer 
$E^{(i)}$ and $\tilde E$.
However, by focusing on $E^{pop}$ and $\tilde E^{(i)}$, as opposed to directly considering 
subject-specific edges, 
a far richer description of functional architecture is obtained. 
% Edges that correspond to $E^{pop}$ can be considered as 
% population edges while edges belonging to $\tilde E^{(i)}$ are 
% variable edges indicating subject-specific idiosyncrasies. 
In the case of the motivating example presented in Section \ref{sec--Intro}, 
$\tilde E  = \tilde E^{(i)} = \{ A\}$ for all subjects while
the remaining edges are captured in $E^{pop}$.
From the perspective of neuroimaging, partitioning edges in 
this manner is fundamental to further understanding  
the functional architecture of the brain \citep{kanai2011structural, zilles2013individual}.

It is useful to note that the model described in equation (\ref{CovEq}) generalizes two typical approaches in the study of
functional connectivity. The traditional method of estimating a single population network, $\Theta^{pop}$,
by concatenating data across all subjects is equivalent to the assumption that
$\tilde E=\emptyset$. This corresponds to the 
sizable assumption that all observations across all subjects share an identical conditional 
independence structure. 
Conversely, the approach of estimating a functional connectivity network for each subject independently
corresponds to the assumption that $E^{pop}=\emptyset$ (or equivalently
that $\Theta^{pop} = \mbox{\textbf{\textit{0}}}_{p\times p}$). In such a scenario, there
is no advantage to be gained by leveraging information across subjects.
Typically, we would expect the true underlying network structure across subjects to 
lie somewhere along the spectrum between these two extremes; 
thus justifying the proposed model. 

\subsection{Mixed Neighborhood Selection}
\label{sec--MNS}
In this section we formally describe the proposed MNS method. %Mixed Neighborhood Selection method. 
The proposed method can be seen as an extension of neighborhood selection to the \textit{replicated} data scenario.
The objective of such an extension is to 
adequately model the inter-subject heterogeneity present 
in fMRI data. 
% Whilst we 
% do not wish to make excessive assumptions such as exchangeability, we wish to exploit 
% shared structure across subjects where appropriate. 

With this in mind we note that the linear regression approach described in 
Section \ref{sec--NghSel} is particularly advantageous in our 
context where we have multiple related (i.e., clustered)
longitudinal observations. Traditionally, such data has been modeled using linear mixed 
effects models \citep{pinheiro2000mixed}. By adequately modeling the 
noise structure of clustered data, linear mixed models are able to effectively separate 
the population (i.e., fixed) effects from the 
subject-specific (i.e., random) effects. In doing so, they provide
a highly flexible tool through which to analyze and describe many diverse datasets.
They are therefore an ideal candidate in our scenario.

We consider learning the neighborhood of node $v \in V$ over a cohort of $N$ subjects. 
We look to extend linear mixed effects models to the neighborhood selection scenario by proposing the following 
linear mixed effects model:
\begin{equation}
 X^{(i)}_v = X^{(i)}_{\backslash v}~ \beta^v + X^{(i)}_{\backslash v}~ \tilde{b}^{(i),v} + \epsilon^{(i), v} ~ \mbox{ for $i=1, \ldots N$ }.
 \label{MNSeq}
\end{equation}
Recall that $X^{(i)}_v$ denotes the time series at node $v$ for subject $i$. 
The model described in equation (\ref{MNSeq}) directly extends
traditional neighborhood selection model by introducing 
random effect terms, $\tilde b^{(i),v}$, for each subject. 
We note that $\beta^v$ corresponds to the shared population neighborhood. 

The random effects are assumed to follow a multivariate 
Gaussian distribution, $\tilde b^{(i),v} \sim \mathcal{N}(0, \Phi^v)$, independently of $\epsilon^{(i), v}$.
Under such a model the the time series at each node thereby follows a Gaussian distribution of 
the following form:
\begin{equation}
 X^{(i)}_v \sim \mathcal{N} \left (X^{(i)}_{\backslash v} \beta^v , X^{(i)}_{\backslash v} \Phi^v \left (X^{(i)}_{\backslash v} \right )^T +\sigma^2 I \right ).
\label{MNSdist}
\end{equation}

The choice of covariance structure for random effects is crucial to both estimating the model 
as well as to 
its interpretability. While it is possible to 
motivate many choices for $\Phi^v \in \mathbb{R}^{p-1 \times p-1}$, in 
this work we limit ourselves to the scenario where $\Phi^v = \sigma^2 \mbox{diag}({\sigma^v}^2)$.
Here $\sigma^v \in \mathbb{R}^{p-1}$ is a vector describing the standard deviation of
the neighborhood of $v$ across the cohort of $N$ subjects. For example, a large $\sigma^v_u$ value is 
indicative of a large variability in the edge between nodes $v$ and $u$. 
% The converse is true for small 
% $\sigma^v_u$. 

For any node $v \in V$, the model 
described in equations (\ref{MNSeq}) and (\ref{MNSdist}) is easily interpretable.
The population, or fixed effects, neighborhood is captured in $\beta^v$. These are the effects that are shared across
the entire cohort of subjects and correspond to the set of edges in $E^{pop}$. 
Meanwhile, the random effects are able to 
capture subject-specific deviations from the population neighborhood and can thereby be employed to 
obtain a network for each subject. 
Formally, the random effects captured in $\sigma^v$ correspond to the 
set of highly variable edges, $\tilde E$.
% Moreover, the random effects also provide valuable information regarding 
% the variability of given edges across the population. 
Finally, we are also able to obtain estimates of $\tilde b^{(i),v}$, which can be employed to 
obtain subject-specific networks. These values correspond to the subject-specific idiosyncrasies,
$\tilde E^{(i)}$.

\subsection{Estimation algorithm}
\label{sec--EMalgo}

The model described in equation (\ref{MNSeq})
contains the following parameters, $ \phi^v = (\beta^v, \sigma^v, \sigma^2) \in \mathbb{R}^{2(p-1)+1}$, which 
must be estimated. Given $\phi^v$ we can subsequently obtain the
best linear unbiased predictions (BLUPs) for each of the random effects, $\tilde b^{(i),v}$,
across subjects \citep{pinheiro2000mixed}.
In this work $\phi^v$ is 
estimated in a maximum likelihood framework. 
Following from 
equation (\ref{MNSdist}), the negative log-likelihood for node $v$ is proportional to:
\begin{equation}
 \mathcal{L}(\phi^v) = \sum_{i=1}^N \frac{1}{2} \mbox{log det}  ~V_v^{(i)} + \frac{1}{2} \left (X_v^{(i)} - X_{\backslash v}^{(i)} \beta^v\right )^T {V_v^{(i)}}^{-1} \left (X_v^{(i)} - X_{\backslash v}^{(i)} \beta^v\right ),
 \label{MNS_EM_incompleteLL}
\end{equation}
where we define $V_v^{(i)}$ to be the variance structure for node $v$ at subject $i$:
\begin{equation}
 V_v^{(i)} =  \sigma^2 \left (X^{(i)}_{\backslash v} \mbox{diag}({\sigma^v}^2) \left (X^{(i)}_{\backslash v} \right )^T + I \right).
\end{equation}

In order to simplify future discussion, 
we re-parameterize the 
random effects component of the mixed effect model described in equation (\ref{MNSeq}). 
We re-define the random effects term as follows:
\begin{equation}
 \tilde b^{(i),v} = \mbox{diag}(\sigma^v) ~b^{(i),v},
 \label{MNS_reParam}
\end{equation}
where $b^{(i),v} \sim \mathcal{N}(0, \sigma^2 I)$. Both the left and right sides 
of equation (\ref{MNS_reParam}) follow the same distribution, however, this will simplify discussion
of the optimization algorithm below.

While it would be possible to maximize equation (\ref{MNS_EM_incompleteLL}) by obtaining the profile likelihood,
in this work we treat the random effects as latent variables and
employ an EM algorithm \citep{mclachlan2007algorithm}. 
Fitting linear mixed effects models in this manner is a popular approach first 
posited by \cite{dempster1977maximum} and for which many efficient algorithms have been proposed \citep{meng1998fast}.
In the context of this work, such an approach will prove 
beneficial when regularization constraints are enforced on some elements of $\phi^v$.
Assuming the random effects, $ b^{(i),v}$, are
observed the complete data log-likelihood is proportional to:
\begin{equation}
\footnotesize
 \mathcal{L}_c(\phi^v) = \sum_{i=1}^N \frac{n+p}{2} \mbox{log} ~\sigma^2 + \frac{1}{2 \sigma^2} \left ( \left |\left | X^{(i)}_v - X^{(i)}_{\backslash v} \beta^v - X^{(i)}_{\backslash v} \mbox{diag}(\sigma^v) b^{(i),v}  \right |\right |_2^2 + {b^{(i),v}}^T b^{(i),v} \right ).
 \label{MNS_EM_completeLL}
\end{equation}

It is important to note from equation (\ref{MNS_EM_completeLL}) that 
the number of parameters remains fixed even as the number of subjects, $N$, increases. 
Therefore
if a sufficiently large cohort of subjects is studied it is possible to estimate all parameters 
without the need to introduce regularization. 
However, regularization is introduced in this work for two reasons.
First, sparse
solutions remain feasible when only a reduced number 
of observations or subjects are available. Second, parsimonious solutions remain easily interpretable 
even in the presence of many nodes.
As a result, we impose an $\ell_1$ penalty on both the fixed as well as random effects. In 
terms of the random effects we penalize the variance terms, $\sigma^v$. If a variance is shrunk 
to zero, the resulting random effect is effectively removed from the model.
The introduction of sparsity inducing penalties yields the following penalized complete-data log-likelihood:
\begin{equation}
 \mathcal{L}^{\lambda_1, \lambda_2}_c(\phi^v)  = \mathcal{L}_c(\phi^v) + \lambda_1 || \beta^v||_1 + \lambda_2 || \sigma^v||_1,
 \label{MNS_EM_PenCompleteLL}
\end{equation}
where $\lambda_1$ and $\lambda_2$ are regularization parameters.
Sparsity at the population level is enforced by $\lambda_1$, while $\lambda_2$ encourages
sparsity in the random effects by shrinking the standard deviation terms, $\sigma^v$. 

An EM algorithm is employed to minimize
% Following \cite{bondell2010joint} we employ an EM algorithm to minimize 
the penalized conditional log-likelihood.
This involves iteratively computing the conditional expectation of latent variables, $Q(\phi; \phi^v)$,
in our case the random effects,
and minimizing the expected conditional log-likelihood with respect to 
parameters $\phi^v$. 

The expectation step (E-step) can be computed in closed form as follows:
\begin{equation}
 b^{(i),v} = \left ( \mbox{diag}(\sigma^v)  {X^{(i)}_{\backslash v}}^T { X^{(i)}_{\backslash v} } \mbox{diag}(\sigma^v) + I \right )^{-1} { X^{(i)}_{\backslash v} }^T \mbox{diag}(\sigma^v) \left ( X^{(i)}_v - X_{\backslash v}^{(i)} \beta^v \right )
\label{MNS_Estep}
\end{equation}
independently
for each subject $i=1,\ldots, N$. It is clear from equation (\ref{MNS_Estep}) that 
if $\sigma^v_u$ is shrunk to zero then the $u$th entry of $b^{(i),v}$ will also  be zero for all subjects.
% Thus the shrinkage penalty on the random effect variance is capable of 
% pruning the random effects across all subjects.

In the minimization step (M-step) the latent variables, $b^{(i),v}$, are assumed 
to be observed. We therefore learn $(\beta^v, \sigma^v)$ by solving the following
convex problem:
% \begin{align}
%  (\beta^v, \sigma^v) &= \underset{\phi^v}{\operatorname{argmin}} \left \{ Q (\phi; \phi^v) \right \} \\
%  &= \underset{(\beta^v\in \mathbb{R}^p, \sigma^v \in \mathbb{R}_+^{p}) }{\operatorname{argmin}} \left \{  \left |\left | X^{(i)}_v - X^{(i)}_{\backslash v} \beta^v - X^{(i)}_{\backslash v} \mbox{diag}(b^{(i),v}) \sigma^v   \right |\right |_2^2 +\lambda_1 ||\beta^v||_1 + \lambda_2 || \sigma^v||_1 \right \}.
% \label{MNS_Mstep}
% \end{align}
\begin{equation}
\footnotesize
  (\beta^v, \sigma^v) = \underset{(\beta^v\in \mathbb{R}^p, \sigma^v \in \mathbb{R}_+^{p}) }{\operatorname{argmin}} \left \{  \left |\left | X^{(i)}_v - X^{(i)}_{\backslash v} \beta^v - X^{(i)}_{\backslash v} \mbox{diag}(b^{(i),v}) \sigma^v   \right |\right |_2^2 +\lambda_1 ||\beta^v||_1 + \lambda_2 || \sigma^v||_1 \right \}.
\label{MNS_Mstep}
\end{equation}
We note that equation (\ref{MNS_Mstep}) is Lasso problem with distinct regularization
parameters applied to the 
fixed and random effects components respectively. 
A vast range of efficient algorithms can be employed to solve 
equation (\ref{MNS_Mstep}). In this 
work gradient descent algorithms \citep{friedman2007pathwise} were employed.
The motivation behind this choice was that 
due to the iterative nature of the EM algorithm employed in this work, 
a lasso problem must be solved at each iteration.
It follows that while solutions from one iteration to the next will typically not 
be identical they will be relatively similar. As a result, considerable computational
gains can be obtained by using past solutions as warm-starts. Gradient descent algorithms
are particularly well-suited for such tasks.

The proposed EM algorithm therefore iterates between equations (\ref{MNS_Estep})
and (\ref{MNS_Mstep}). This is performed iteratively for each node $v \in V$.
Once this is complete, networks can be obtained by applying the AND/OR rules
described in equation (\ref{ANDOR}). The {MNS} procedure is described in Algorithm \ref{algo_MNS}.

 \begin{algorithm}[h!]
 \DontPrintSemicolon
 \KwIn{Data across $N$ subjects, $\{X^{(i)}\}$, regularization parameters, $\lambda_1, \lambda_2$.}
%  \KwResult{}
%  initialization\;
 \Begin{
 \For{$v\in \{1, \ldots, V\}$}{
 Define initial estimates: $\beta^v = \mbox{\textbf{\textit{0}}}$, $\sigma^v=\mbox{\textbf{\textit{1}}}$, $\sigma=1$ and $b^{(i),v}=\mbox{\textbf{\textit{0}}}$
 
 \While{$\mbox{not converged}$}{
 Update $(\beta^v, \sigma^v)$ by solving equation (\ref{MNS_Mstep}) \tcp*{M-step}
 Estimate latent variables using equation (\ref{MNS_Estep}) for $i \in \{1, \ldots, N\}$ \tcp*{E-step}
 }
 Store $\beta^v$, $\sigma^v$ and $\left \{ b^{(i),v} \right \}_{i=1}^N$
 }
 $E^{pop} = \{ (u,v): \beta^v_{u} \neq 0 \mbox{ and } \beta^u_v\neq 0 \} $
 
 $\tilde E = \{ (u,v): \sigma^v_{u} \neq 0 \mbox{ and } \sigma^u_{v}\neq 0 \} $
 
 $\tilde E^{(i)} = \{ (u,v): b^{(i),v}_{u} \neq 0 \mbox{ and } b^{(i),u}_{v}\neq 0 \} $
 }
 \Return{$E^{pop}$, $\tilde E$ and $\tilde E^{(i)}$ for $i=1, \ldots, N$}
 \caption{{Mixed Neighborhood Selection}}
 \label{algo_MNS}
\end{algorithm}

\subsection{Parameter tuning}
\label{sec--ParamSelect}

The proposed method requires the tuning of two 
regularization parameters which govern the 
nature of the estimated population and subject-specific networks respectively. 
Large values of $\lambda_1$ will lead to 
sparse networks at the population level. Conversely, selecting 
large $\lambda_2$ will penalize the variance 
of the random effects leading to 
sparse subject-specific contributions to 
covariance structure.

Moreover, in the class of models considered in this work each 
covariate can contribute to the 
fixed as well as random effect structure. % (or both). 
This can potentially lead to problems regarding the interpretability of
estimated models.
For example, over-penalizing the 
fixed effects may lead to over-estimation of the 
random effect variances as compensation \citep{schelldorfer2011estimation}. 
The choice of regularization 
parameters is therefore a delicate issue which must be handled with care. 

While information theoretic methods such as the Bayesian Information Criterion (BIC)
may be employed for the purpose of tuning regularization parameters,
in this work we employ cross-validation (CV) \citep{arlot2010survey}.
As such, the data is divided into $K$ folds.
For each fold, the data from the remaining $K-1$ folds is employed to 
fit the penalized linear mixed model described in Section \ref{sec--MNS}. 
The resulting model is them used to 
predict the unseen data and the mean-square error is noted.
This procedure is repeated and the results are averaged across all nodes. The 
pair of parameters %$(\lambda_1, \lambda_2)$ 
which minimize the mean-square error are subsequently selected. 

% \subsubsection{Computational complexity and number of effective parameters}
% 
% As noted previously, the proposed method possess several computationally 
% appealing properties. The first of these is the fact that it can be 
% trivially parallelized across nodes. %, leading to considerable improvements in running time.
% Moreover, the EM algorithm described in Section \ref{sec--EMalgo} is also computationally
% efficient. The M-step involves solving a lasso problem with $2(p-1)$ regressors and 
% $Nn$ observations, resulting in a computational complexity 
% of $\mathcal{O}(Nnp\min \{Nn, p \})$ \citep{efron2004least}. Moreover, efficiently solving 
% lasso problems has received considerable attention in recent years and has lead to 
% the discovery of many highly efficient algorithms \citep{beck2009fast}.
% The E-step involves inverting a $(p-1) \times (p-1)$ matrix for each subject, leading to a
% computational complexity of $\mathcal{O}(Np^3)$. As a result, the proposed method has a complexity of
% $\mathcal{O}(Nnp^2\min \{Nn, p \} + Np^4)$.
% 
% An additional advantage of the proposed method is that the number of parameters in 
% the model remains fixed as the number of subjects increases. This is in 
% contrast to alternative methods where the number of parameters to estimate
% increases with every subject. 
% This property is a direct result of the mixed effects component 
% introduced in Section \ref{sec--MNSmodel}. 
% This makes the proposed method particularly suitable to 
% datasets containing a large number of subjects with a low signal-to-noise ratio.

\subsection{Relationship to previous work}
\label{sec--RelWork}

The problem of simultaneously estimating multiple graphical models has recently received 
considerable attention. This problem is particularly relevant in the context of neuroscience
where multi-subject studies are commonplace. Moreover, the recent
collaborative trend towards sharing large neuroimaging datasets serves 
as an additional motivation \citep{poldrack2014making}. 

From a methodological perspective several solutions have been proposed. 
\cite{varoquaux2010brain}
propose to leverage information by imposing a common sparse support for the precision matrices 
of all subjects. This is achieved via the use of a separate group Lasso penalty on the each edge. 
A similar penalty is also employed by \cite{wee2014group}
in the context of performing classification between normal controls and MCI subjects.
The Joint Graphical Lasso (JGL), proposed by \cite{danaher2014joint}, generalized the above methods. Under their 
proposed framework, any convex penalty can be applied on edges across subjects and they propose both group 
as well as a fused lasso penalties.
Other related methods include 
the work of \cite{guo2011joint} and \cite{pierson2015sharing}. 

Our proposed method holds a number of advantages over these preceding methods. First, 
it provides a principled manner for uncovering a population network that reflects edges shared across 
all subjects. While this could be achieved by combining observations across all subjects, 
such an approach carries with it
the tenuous assumption that observations are interchangeable across subjects and will
almost certainly be violated in practice. Approaches such as the JGL readily obtain estimates for each subject, however,
there is no clear method for obtaining a population network. Second, the proposed method 
also reports edges which demonstrate high variability across a cohort. As far as we are aware,
this is the first method to achieve this. Third,  as we demonstrate in the simulation study below, 
the proposed method is capable of accurately recovering networks on both a subject and population level.
We attribute this to the covariance model described in Section \ref{sec--CovModel}; by 
accurately learning which edges correspond to population (i.e., fixed effects) edges and 
which correspond to subject-specific idiosyncrasies the proposed method is able to
leverage information across subjects in a discriminative manner. 
Lastly, the {MNS} algorithm is easily amenable to 
parallelization, making it particularly suitable for large datasets.

However, we must also note some of the disadvantages of our approach. The most significant of which is that 
we require an EM algorithm to estimate our model and therefore do not enjoy the guarantees 
provided by the convexity of the previous methods such as the JGL.
In addition to this, since we take a neighborhood selection 
approach the resulting network estimates are
not precision matrices and are therefore not guaranteed to be positive definite. 

Finally, a related problem is that of quantifying statistically significant differences across multiple populations. 
\cite{narayan2015two} provide a principled manner for uncovering such differences. 
Much like the our work, they incorporate random effects into their model; however they do so in the context 
of modeling the presence or absence of an edge in multiple bootstrap 
estimates of the connectivity networks. In this way they are able to 
perform hypothesis tests on an edge-by-edge basis \citep{narayan2013randomized,narayan2015two}. 
Multiple other methods have also been developed for quantifying differences 
across populations. \cite{ginestet2014hypothesis} develop a central limit theorem for 
graph laplacians, allowing them to derive pivotal quantities and 
formally test for differences in pairs of networks. \cite{zalesky2010network} propose a network-based
statistic to control the family-wise error rate when performing mass-univariate testing across
all edges.

\section{Simulation Study}
\label{sec--SimStudy}

In this section we evaluate the performance of the proposed method using simulated data
that is representative of real functional imaging data.
We assess the empirical performance of the {MNS} algorithm in three distinct settings which correspond 
to correctly reporting the edge structure of the population, subject-specific and 
highly variable network edges respectively.
The first task corresponds to correctly recovering $E^{pop}$ while the second 
requires learning subject-specific edge structure, $E^{(i)}$, defined in 
equation (\ref{SubjectEdgeSupport}). Finally, the task of recovering variable edges
is equivalent to learning the set of variable edges, $\tilde E$.

\subsection{Network simulation}

In order to perform such a study we require a method through which to simulate population 
and subject-specific networks. 
While numerous algorithms have been proposed to generate 
random individual networks
(e.g., the \cite{erdHos1959random} model and the preferential
attachment model of \cite{barabasi1999emergence}), there has been limited work on algorithms to simulate multiple
related networks.
Notably, there is no documented method through which to generate networks from a cohort of 
related subjects that demonstrate the characteristics observed in real fMRI data; namely a shared core
structure which is reproducible across all subjects together with significant inter-subject variability
in the remaining edges \citep{bullmore2009complex, ashby2011statistical, poldrack2011handbook}. 

In order to address this issue we propose a new method of simulating networks. %, outlined in Algorithm \ref{algo1}.
The proposed algorithm is motivated by an exploratory data analysis of resting state fMRI data from healthy
subjects within the ABIDE consortium dataset.
We briefly outline the proposed algorithm in this section with further details 
provided in Appendix \ref{app--Simulating}.

The proposed network simulation method proceeds as follows: first a population edge set, corresponding to $E^{pop}$,
is randomly sampled according to the preferential attachment model of \cite{barabasi1999emergence}. 
These edges constitute the core, reproducible connectivity structure which will be present across all subjects. 
The precision matrix, $\Theta^{pop}$, corresponding to $E^{pop}$ is then obtained by 
randomly sampling edge weights as described in \cite{danaher2014joint}.

A set of variable edges, $\tilde E$, is then selected randomly according to the \cite{erdHos1959random} model, 
possibly overlapping with $E^{pop}$. 
For each subject, edges in $\tilde E$ are included in the subject-specific network %(i.e., in $\tilde E^{(i)}$)
with some fixed probability $\tau \in [0,1]$. 
The edges weights associated with $\tilde E^{(i)}$ are randomly sampled for each subject and stored in $\Theta^{(i)}$. 
We note that the special case where $\tau=0$ yields an identical
network for all subjects. On the other hand, setting $\tau=1$ still results in inter-subject variability
as all edges within $\tilde E$ will have varying weights across subjects. 
This method is summarized in Algorithm \ref{algo1}.

 \begin{algorithm}[h!]
 \DontPrintSemicolon
 \KwIn{Number of nodes $p$, number of subjects $N$, size of random effects network $e_{ran} = |\tilde E|$,
 a random effects edge probability $\tau \in [0,1]$ and connectivity strength $r \in \mathbb{R}_{+}$}
%  \KwResult{Population network, $\Theta^{pop}$, subject-specific networks, $\{\Theta^{(i)}\}$, random effects edges
% $\tilde E$}
 \Begin{
 Simulate $E^{pop}$ according to \cite{barabasi1999emergence} model\;
 Build $\Theta^{pop}$ by uniformly sampling edge weights from the 
%  by randomly 
%  selecting edge weights from the
 interval $[-r, -\frac{r}{2}] \cup [\frac{r}{2}, r]$ \;
 Simulate $\tilde E$ according to \cite{erdHos1959random} model with $e_{ran}$ edges\;
 \For{i $\in \{1, \ldots, N\}$}{
 \For{each edge $(j,k)$}{
 \If{$(j,k) \in \tilde E$}{
%  $(j,k) \in E^{(i)}$ with probability $\tau$
  $E^{(i)} \leftarrow E^{(i)} \cup (j,k) $ with probability $\tau$
 }
%  \Else{pass}
 }
%  Define $\Theta^{(i)}$ as follows: 
%  $$ \Theta^{(i)}_{k,j} = \Theta^{(i)}_{j,k} =  \begin{cases}
%   0, & \text{if }  (k,j) \notin \tilde E \\ %\Theta^s_{j,k} = 0, \\
%   1, & \text{with probability } \tau.
% \end{cases} ~~ \mbox{for}~1\leq j<k\leq p $$
Randomly select edge weights and signs for $\Theta^{(i)}$
 }
 }
 \Return{$E^{pop}, \tilde E$, $\{ E^{(i)}\}$ and $\Theta^{pop}$, $\{\Theta^{(i)} \}$}
 \caption{Generate population and subject-specific random networks}
 \label{algo1}
\end{algorithm}

Algorithm \ref{algo1} was employed to simulate synthetic data for a cohort of $N=10$ subjects.
The number of nodes was fixed at $p=50$, the size of the random
effect network  at $e_{ran}=20$ and the random edge probability
was $\tau=1$.
Datasets for each 
subject were simulated according to:
% $n=200$ observations where simulated per subject as 
% follows:
\begin{equation}
 X^{(i)} \sim \mathcal{N} \left ( 0, \left (  {PD} \left(\Theta^{pop} + \Theta^{(i)} \right )\right )^{-1} \right ),
\end{equation}
where $PD(\cdot)$ is a function applied in order to ensure the resulting matrix is positive definite 
(see Appendix \ref{app--Simulating}).
Data was simulated with varying numbers of observations per subject, $n \in \{50,100,200\}$.
We note that in all cases the number of observations per subject is far below the 
amount that would be sufficient to obtain a maximum likelihood estimate of the precision matrix. As a 
result, it is crucial to effectively leverage information across subjects.

Simulating networks as described in Algorithm \ref{algo1} is only one of many 
possible methods which could be employed. In order
to provide a thorough and fair comparison an additional simulation was also 
performed where networks were simulated  
as described in \cite{danaher2014joint}. 
This simulation was proposed with the objective of providing empirical evidence regarding 
how accurately subject-specific networks could be reported. It is therefore not well suited 
for examining how reliably the population or variance networks can be reported. 
The results for this simulation are provided in Appendix \ref{DanaherSims}.

\subsection{Alternative models}
Throughout this simulation the performance of the {MNS} algorithm 
was benchmarked against the current state of the art in each of 
the three settings described above. In
the case of estimating the population network, the \textit{graphical lasso} (Glasso) \citep{friedman2008sparse}
was employed. Such an approach has been used extensively in the neuroimaging community to learn
functional connectivity networks across populations \citep{smith2011network, varoquaux2013learning}.
Moreover, an approach based on  resampling and randomization was also employed.
This 
approach, which we refer to as the \textit{Stability} approach, is outlined in Appendix \ref{app--Stability}. 
We note that while this approach is inspired by the recently proposed $R^3$ method of \cite{narayan2015two},
the objective here is different. Formally, the $R^3$ method %developed by \cite{narayan2015two}
% specifically 
addresses the issue of comparing covariance structure across two populations while
this simulation study is based on only a single population. 

The problem of estimating subject-specific functional connectivity networks has
received considerable attention. In this simulation study we compare the
performance of the proposed method with the two penalized likelihood methods presented in 
\cite{varoquaux2010brain} and \cite{danaher2014joint}. Both methods are related, enforcing a
group or a fused lasso penalty across edges respectively, and we refer to each as 
the {JGL-Group} or the {JGL-Fused} algorithms respectively. The 
Glasso algorithm is also employed in this context. 

As far as we are aware there are no alternative methods available 
which address the problem of recovering highly variable edges.
In order to provide a benchmark for the {MNS} algorithm, the aforementioned \textit{Stability} approach 
was employed in this context.

\subsection{Performance measures}

Throughout this simulation the task of recovering covariance structure is treated as a binary classification 
task. Thus performance is measured according to the proportion
of edges which are correctly reported as being either present or absent. 
In order to compare performance across various algorithms we employ receiver operating characteristic
(ROC) curves, which illustrate the performance of a binary classifier
by plotting the true positive rate against false positive rate across a 
range of regularization parameters \citep{krzanowski2009roc}. 

The use of ROC curves requires a single, sparsity-inducing parameter to be varied across a range of 
possible values.
As discussed previously, in the case of the {MNS} algorithm 
both the population and subject-specific parameters can affect 
sparsity.
% \footnote{Consider the example where a very small value for $\lambda_2$ is employed relative to $\lambda_1$. 
% Here the random effects will also be inflated to also account for fixed population terms which will
% be pruned.}. 
As a result, we look to reparameterize the {MNS} penalty as follows:
\begin{align}
 \label{MNS_reparam1}
 \lambda_1 &= \alpha \lambda\\
 \label{MNS_reparam2}
 \lambda_2 &= \sqrt{2} (1-\alpha)\lambda
\end{align}
where $\alpha$ controls the ratio of sparsity between the population and subject-specific contributions
and $\lambda$ the overall sparsity. Thus $\alpha$ is fixed allowing $\lambda$ to vary.
While no such adjustments are needed in the case of the {JGL-Fused} algorithm, we follow the
same parameterization described in 
equations (\ref{MNS_reparam1}) and (\ref{MNS_reparam2}) in the case of 
the {JGL-Group} algorithm\footnote{Note this 
same parameterization was employed in the original study described in \cite{danaher2014joint}}.%:

\subsection{Simulation results}
In this section we present the results to the simulation study described above. 
We begin by first considering performance in the context of recovering the 
set of variable edges. This is a fundamental problem in neuroscientific applications \citep{kelly2012characterizing}, 
however to date it has 
received limited attention. 
Results for the more frequently studied problems of  recovering population and subject covariance structure 
are presented in Section \ref{Res--Pop} and \ref{Res--Sub} respectively. 

Throughout this simulation the {MNS} algorithm was run with $\alpha = 0.25$ while sparsity parameter $\lambda$ varied as
described in equations (\ref{MNS_reparam1}) and (\ref{MNS_reparam2}). 
The same parameterization was employed for the {JGL-Group} algorithm with $\alpha=0.15$ selected.
In the case of the {JGL-Fused} algorithm, $\lambda_2=0.2$ 
was employed. Finally, the \textit{Stability} algorithm was run with 
$B=10,000$ bootstrap iterations per subject and $c=0.25$.

  \begin{figure}[h!]
  \centering
\includegraphics[width=\textwidth]{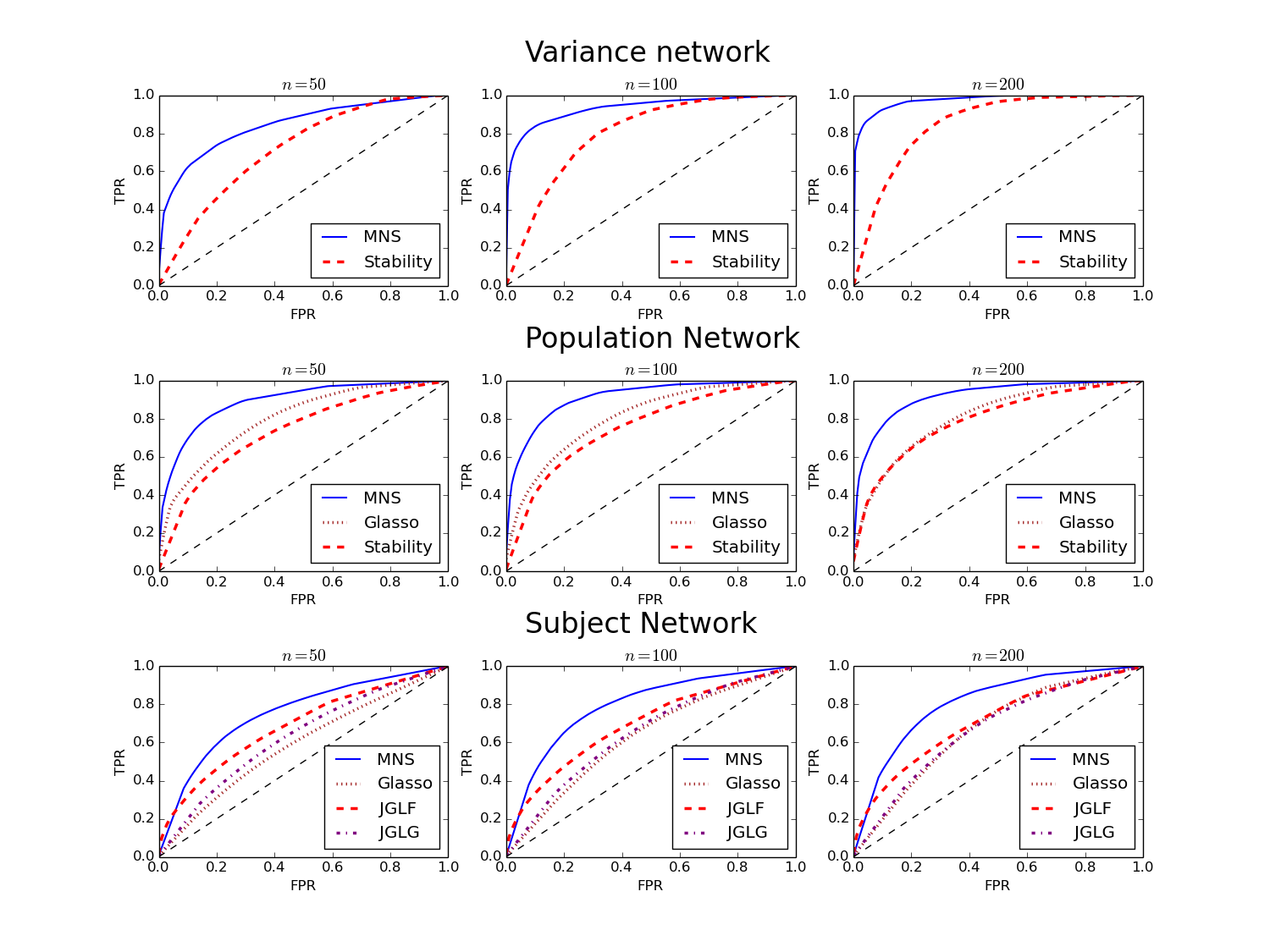}
\caption{Simulation results for all five algorithms
across all tasks. Recovery of variable edges is considered in the top panel,
population network recovery in shown in the middle panel and finally
the bottom panel shows subject-specific network recovery.
This simulation was performed with $p=50$ nodes and $n\in\{50,100,200\}$ observations.}
\label{SimResultsFig}
\end{figure}

\subsubsection{Variable network recovery}
\label{Res--Var}
Understanding variability in covariance structure across a cohort of subjects is a fundamental
problem in neuroscience. 
In particular, understanding whether this variation can be attributed to 
phenotypic characteristics or other sources of noise (e.g., physiological noise or scanner noise)
is crucial in further understanding the architecture of the human connectome. 

The results shown in the top panel of Figure [\ref{SimResultsFig}] demonstrate 
that the proposed {MNS} algorithm is able to accurately identify 
edges which demonstrate variability across a cohort of subjects. 
This is in contrast to the \textit{Stability} method. % employed whose performance is considerably worse.
Briefly, the \textit{Stability} method (described in further detail in Appendix \ref{app--Stability})
treats the presence or absence of edges at a subject level as a Bernoulli random variable. 
A hierarchical random effects model is then proposed to leverage information across all subjects. 
The resulting estimate of the edge variability, given in equation (\ref{rhopop}), 
is then employed to discriminate between variable and non-variable edges.
The \textit{Stability} method therefore
corresponds to a two-step procedure where information is only leveraged \textit{after}
networks have been estimated for subjects independently. 
This is in contrast to the proposed method where subject-specific, population and variable networks are
learnt \textit{simultaneously}. This allows the {MNS} algorithm to effectively leverage information
and results in significant improvements in performance. % shown in Figure [\ref{SimResultsVar}].
Further results are given in Table [\ref{ResultsTable}] where the true positive rate (TPR) and false
positive rate (FPR) are reported for  selected regularization parameters. 

\subsubsection{Population network recovery}
\label{Res--Pop}

Obtaining an accurate understanding of a population level covariance structure is a 
challenging problem due to the high inter-subject variability. 
As mentioned previously, it is imperative to differentiate between 
subject-specific idiosyncrasies and behavior which is reproducible across 
the entire cohort. 
A popular approach often taken in neuroimaging 
studies is to estimate a single network using data from all subjects, thus effectively concatenating 
individuals' data. This corresponds to the sizable  assumption that $\tilde E = \emptyset$. This approach
is included in this simulation together with 
the aforementioned \textit{Stability} approach.
Here the mean estimate for edge presence, given by equation (\ref{mupop}) was employed.

Results are shown in the middle panel of Figure [\ref{SimResultsFig}]. 
It is interesting to note that for small sample sizes (i.e., $n=50$ or $n=100$)
the \textit{Stability} approach is out-performed by the 
Glasso. 
As mentioned in Section \ref{Res--Var},
we attribute this drop performance to 
its two-step design whereby information is only leveraged 
\textit{after} networks have been estimated. It is only when the number of observations
increases that reliable estimates of uncertainty can be obtained. 
Conversely, the difference in performance between the Glasso algorithm and the 
{MNS} algorithm is due to the presence of variable edges, implying $\tilde E \neq \emptyset$. 
Thus, by providing a more sophisticated model for inter-subject variability, the {MNS} algorithm
is able to obtain more reliable population network estimates.

\subsubsection{Subject-specific network recovery}
\label{Res--Sub}
Finally, we consider recovering subject-specific networks.
This problem has received considerable attention in recent years and a range of methods have been 
proposed. The underlying theme in these methods revolves around effectively leveraging information 
across subjects. This is particularly evident in the JGL algorithm proposed by \cite{danaher2014joint}
and the method presented by \cite{varoquaux2010brain}. However, a short-coming of the 
aforementioned methods is that they leverage information in an indiscriminate manner.
By enforcing either a group or fused lasso penalty on the 
entries of precision matrices, such methods effectively encourage homogeneous information leveraging 
across all edges.
% in an 
% indiscriminate manner. %This makes such approaches susceptible to 

While such approaches constitute significant methodological 
improvements, we envisage a scenario where edges can be ordered according to their variability (or reproducibility).
This is a well-documented phenomenon in neuroimaging. 
In particular for fMRI data 
there is compelling evidence to suggest that variability in functional connectivity is directly modulated 
by the distance between regions \citep{power2012spurious, satterthwaite2012impact,van2012influence}.

The proposed {MNS} algorithm is able to address precisely this issue. By discriminating 
between subject-specific and population edges, it is able to effectively vary the 
how extensively each edge is leveraged across a population. As a result, the {MNS} algorithm 
is able to more reliably recover subject-specific covariance structure, as shown 
in the bottom panel of Figure [\ref{SimResultsFig}].

\begin{table}[ht]
\centering
\begin{tabular}{c|c|cc|cc|cc}
  \hline
Algorithm & $n$ & \multicolumn{2}{c}{Population} & \multicolumn{2}{c}{Subject} & \multicolumn{2}{c}{Variance}\\ 
  \hline
  \hline 
  & & TPR & FPR & TPR & FPR & TPR & FPR\\
  \hline
  \multirow{3}{*}{MNS} 
    & 50 & 0.76 & 0.12 & 0.75 & 0.33 & 0.54 & 0.06 \\ 
   & 100 & 0.77 & 0.11 & 0.80 & 0.32 & 0.70 & 0.03 \\ 
   & 200 & 0.75 & 0.11 & 0.82 & 0.30 & 0.79 & 0.02 \\ 
   \hline
  \multirow{3}{*}{Glasso}  
  &  50 & 0.69 & 0.27 & 0.88 & 0.83 & \multicolumn{2}{c}{\multirow{3}{*}{NA}}  \\ 
   & 100 & 0.70 & 0.27 & 0.83 & 0.66 &  &  \\ 
   & 200 & 0.68 & 0.27 & 0.85 & 0.58 &  &  \\ 
   \hline
   \multirow{3}{*}{Stability}  
& 50 & 0.56 & 0.20 & \multicolumn{2}{c|}{{\multirow{3}{*}{NA}}} & 0.54 & 0.24 \\ 
   & 100 & 0.59 & 0.20 &  &  & 0.64 & 0.18 \\ 
  & 200 & 0.78 & 0.35 &  &  & 0.71 & 0.15 \\ 
  \hline
\multirow{3}{*}{JGL Group}
 &50 &  \multicolumn{2}{c|}{{\multirow{3}{*}{NA}}}  & 0.86 & 0.71 &  \multicolumn{2}{c}{{\multirow{3}{*}{NA}}}  \\ 
   & 100 &  &  & 0.83 & 0.62 &  &  \\ 
  & 200 &  &  & 0.82 & 0.57 &  &  \\ 
  \hline 
  \multirow{3}{*}{JGL Fused} &
  50 &   \multicolumn{2}{c|}{{\multirow{3}{*}{NA}}}   & 0.78 & 0.51 &   \multicolumn{2}{c}{{\multirow{3}{*}{NA}}}  \\ 
   & 100 &  &  & 0.79 & 0.51 &  &  \\ 
   & 200 &  &  & 0.79 & 0.50 &  &  \\   
   \hline
\end{tabular}
\caption{Performance of all five algorithms considered. Results are presented 
for each of the three tasks: recovering population, subject and variance networks. 
For each algorithm the true positive rate (TPR) and false positive rate (FPR) are reported where 
applicable.}
\label{ResultsTable}
\end{table}

\section{Application}
\label{sec--APP}
In this section the proposed {MNS} algorithm is applied to 
resting-state fMRI data from 
the ABIDE dataset \citep{di2014autism}.
While the ABIDE dataset contains data corresponding to healthy subjects and autism 
spectrum disorder (ASD) subjects, we chose only to study healthy controls here
as the focus of this work consisted in fully understanding uncertainty across a single population of subjects. 
The decision to study the ABIDE dataset in this manner was motivated by the 
fact that it is an open-source dataset which has been widely studied, albeit often in the context
of autism.
% where there is considerable variability
% in the age of subjects. 
While it would have been possible to study resting-state data for healthy subjects across all 
sites, only the data from the 
University of Utah, School of Medicine (USM)
site was considered here.
% This site was selected as it included both children as well as adults (ages ranged from
% 11 to 45 years old). As the objective of the {MNS} algorithm is to report variability 
% the 
This consisted of 58 subjects with ages ranging from 11 to 45 years old. 

\subsection{Data acquisition and processing}

Resting-state fMRI data for healthy subjects was collected from the USM site \citep{di2014autism},
resulting in a cohort of 58 subjects. 
Whole-brain functional images were acquired by a 3T Siemens Magnetom TrioTim
using an EPI sequence (T$2^*$-weighted gradient echo; TR=2000 ms; TE=28 ms; flip angle=$90^\circ$;
40 interleaved slices; voxel size= $3.4\times 3.4 \times 3.0$ mm).
In addition, structural images
of each subject were acquired using a 3D-MPRAGE sequence (T1-weighted gradient echo
TR= 2300 ms; TE = 2.91 ms; TI = 900 ms;
flip angle  = $9^{\circ}$; 160 sagittal slices; matrix size = $240 \times 256$;
FOV = 256 mm; slice thickness = 1.2 mm).
Preprocessing of the 
structural images 
involved removal of background noise and 
segmentation using 
 using FAST \citep{zhang2001segmentation}
and FIRST \citep{patenaude2011bayesian}.
This resulted in 92 anatomical regions of interest.

Preprocessing of functional images involved motion correction using MCFLIRT and 
bandpass filtering (0.01-0.1 Hz) to extract low frequency fluctuations. 
In addition, a spatial single subject ICA was performed using FSL MELODIC \citep{smith2004advances}. 
Resulting independent components were automatically identified as 
either signal or noise using FSL FIX \citep{salimi2014automatic}.  
Functional data was subsequently denoised by regressing out noise independent
components. 
Finally, motion parameters were regressed out. 
The first five volumes of each cleaned scan 
where subsequently discarded in order to 
allow for signal equilibration. 
Mean time-courses were extracted from all 92 regions, resulting
in datasets with 230 observations over 
92 nodes for each subject.
 
% Preprocessing of functional images involved motion correction 
% using MCFLIRT and bandpass filtering (0.01-0.1 Hz) to extract low frequency 
% fluctuations. The first five volumes of each scan were discarded
% in order to allow for signal equilibration. 
% 
% MELODIC was then employed 
% to perform group ICA across all subjects \citep{smith2004advances}.
% Results were then classified as either signal or noise using FIX. 
% Mean time-courses were extracted from all 92 regions, resulting
% in datasets with 230 observations over 
% 92 nodes for each subject.

% Further preprocessing involved 
% motion correction using MCFLIRT and 
% bandpass filtering (0.01-0.1 Hz) to extract low frequency 
% fluctuations. 
% Finally, the first five volumes of functional images where dropped.
% This resulted in a datasets with 230 observations over 92 nodes
% for each subject. 

It follows that the number of observations 
available is far lower than the number required
to
accurately estimate the 
covariance structure for each subject; thereby implying that 
efficient leveraging of information across subjects is crucial.

% Functional preprocessing included dropping 
% the first 4 volumes of functional images, removing the spikes,
% slice timing, and motion correction.
% Band-pass temporal filtering (0.01-0.1Hz) 
% was performed to extract the low frequency fluctuations. 
% Finally, 92 anatomical regions of interest where selected.
% This resulted in a datasets with 230 observations over 92 nodes
% for each subject. It follows that the number of observations 
% available is far lower than how many would be required
% to
% accurately estimate the 
% covariance structure for each subject; thereby implying that 
% efficient leveraging of information across subjects is crucial. 

\subsection{Results}
The {MNS} algorithm requires the specification of two regularization parameters, each
of which controls the population and subject-specific topology of each node. 
As discussed in Section \ref{sec--ParamSelect},
the choice of these parameters is crucial to the interpretation of the resulting model.
Parameter were selected on the basis of a 10-fold cross-validation framework.
% Formally, a two-dimensional extensive grid-search was performed over a range of values for 
% $\lambda_1$ and $\lambda_2$. The resulting choice of regularization parameters was the pair that
% minimized average cross-validation error. These parameters were subsequently used to fit 
% the proposed {MNS} model using the entire dataset. 

One of the advantages of the proposed {MNS} algorithm is that it is able to simultaneously 
estimate both a population network, corresponding to reproducible edges which are present across the entire 
cohort of subjects, as well as a network quantifying variability on an edge-by-edge basis. The latter network 
is able to succinctly summarize variability across a cohort of subjects on an edge-by-edge basis. 
% Both these networks are shown in Figure [\ref{ResultsMNS}].

\begin{figure}[h!]
  \centering
\includegraphics[width=\textwidth]{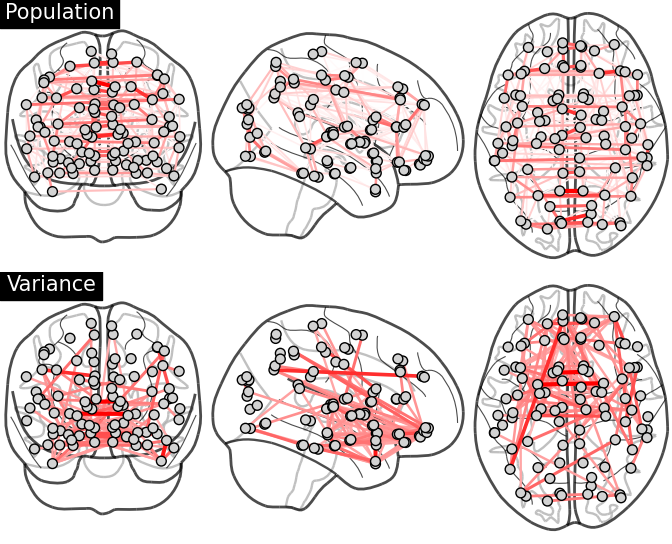}
\caption{Estimated population network (top) and variable edge network (bottom). 
Edges color indicates positive (red) or negative (blue) partial correlations and
thickness is proportional to the magnitude of the partial correlation (or variance in the case of variance network).
In the case of the population network we note there is high inter-hemispheric coupling which is to be expected 
in resting-state data. Conversely, the variable edge network demonstrates interesting spatial structure; 
highly variable edges occur between spatially proximal or remote regions as often hypothesized within 
the neuroimaging community. We also note that
variability seems to conglomerate near the prefrontal and orbitofrontal regions.}
\label{ResultsMNS}
\end{figure}

On the top panel of Figure [\ref{ResultsMNS}], the estimated population network is shown. 
The estimated network has an estimated edge density of around $10\%$ and we note there is strong
inter-hemispheric coupling as would be expected in resting-state connectivity. 
More importantly, the bottom panel of Figure [\ref{ResultsMNS}] shows the estimated variability across all edges where the 
edge thickness is proportional to the estimated variance of the random effect. 
A hallmark of this plot 
% There are two clear hallmarks of the estimated variability network; the first 
is the anatomical distribution of highly variable edges.
It has been suggested within the neuroimaging literature that 
head-movement may lead to the presence of spurious noise structures in fMRI data \citep{van2012influence, satterthwaite2012impact, power2012spurious, patel2014wavelet}.
In particular, it has been suggested that 
subject motion in the scanner results in decreased estimated correlations between spatially remote 
regions and increased estimated correlations across spatially adjacent regions \citep{power2012spurious}.
Our results serve as a corroboration of this hypotheses as we discuss below.

\begin{figure}[h!]
  \centering
\includegraphics[width=.8\textwidth]{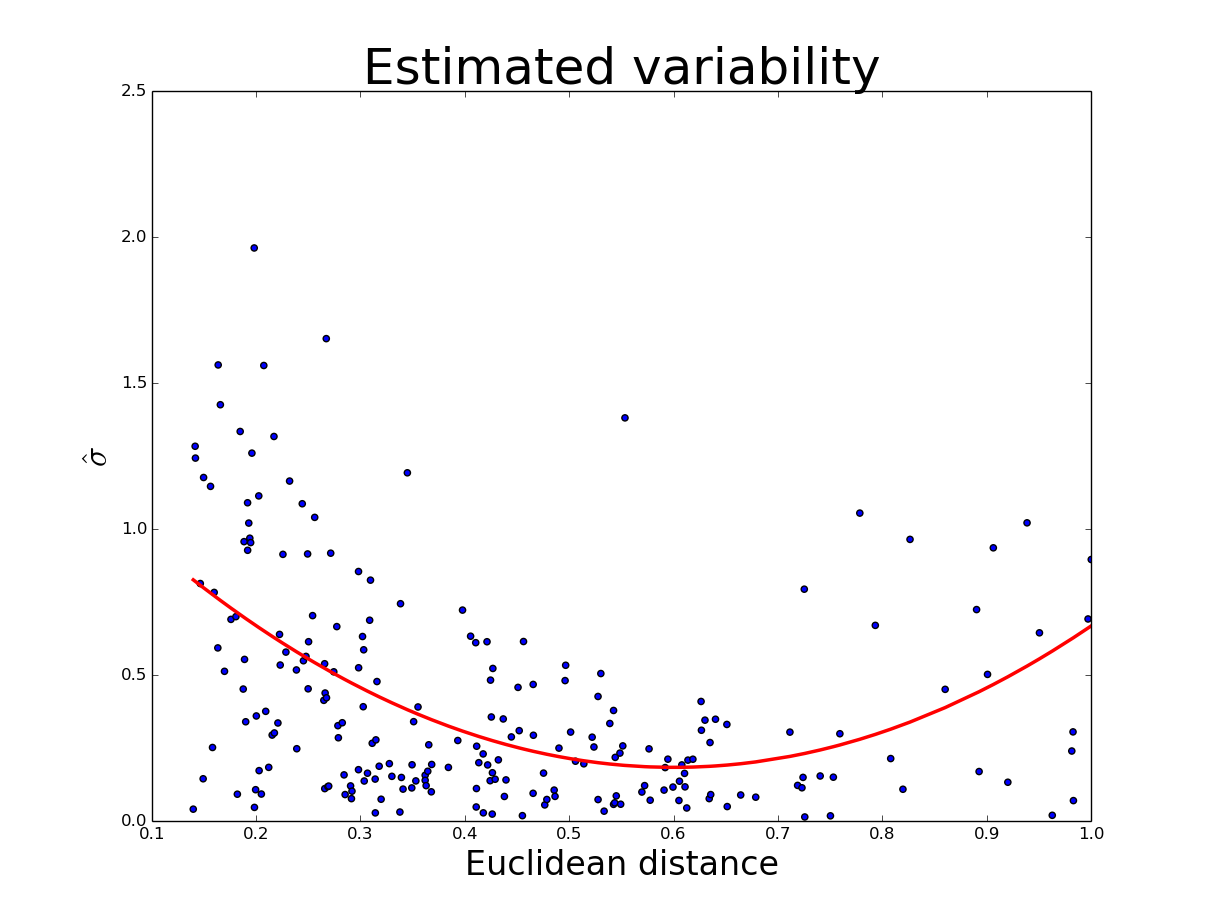}
\caption{Estimated edge variability as a function of (scaled) Euclidean distance between regions. 
The quadratic line of best fit in plotted in red. 
This serves to highlight our claim that variability is highest between spatially adjacent or remote regions. }
\label{DistanceRelation}
\end{figure}

Figure [\ref{DistanceRelation}] plots the estimated variance against euclidean distance for each edge. 
We note there is a quadratic effect, 
with the large variances exhibited between spatially proximal and remote regions. 
% observed to occur at edges between proximal and remote regions. 
This is precisely the effect reported in various neuroimaging studies (e.g., \cite{satterthwaite2012impact}
and \cite{power2012spurious}). Following from the aforementioned works, we 
hypothesize that this variability may have been introduced by the individual head motion of each subject. 

It is also interesting to study the localization of 
highly variable edges in the bottom panel of Figure [\ref{ResultsMNS}]. In particular, we note there is 
high variability within both the 
prefrontal and orbitofrontal regions; this is clearly seen in the sagittal and axial plots. 
These regions are known to be particularly problematic \citep{deichmann2003optimized}
due to their proximity to air/tissue interfaces which leads to image distortion and 
signal losses and have been reported as being highly variable across subjects \citep{finn2015functional, miranda2014connectotyping}.
It is therefore reassuring that the proposed method is able to report such variability
in order to obtain more reliable networks at the population and subject-specific level. 

\section{Discussion}

Understanding variability in functional connectivity
across a population of subjects remains a fundamental challenge
with important consequences in modern neuroimaging. 
Formally, understanding variability is seen as a crucial
stepping stone towards obtaining a more holistic understanding and interpretation of functional
connectivity \citep{kelly2012characterizing,fallani2014graph}. This in turn will naturally lead to 
a better understanding of many diseases which are thought to be caused or 
characterized by changes in functional connectivity \citep{uddin2013reconceptualizing}. 
From a statistical perspective estimating variability in second order 
statistics
remains challenging due to positive-definiteness and identifiability constraints.
% to the fact that such statistics are often constrained
% to be positive semi-definite and 
% the need for models to be identifiable in order to perform inference.
Mixed effects models are ideally suited in the context of  regression \citep{pinheiro2000mixed}, 
and have been extend to the domain of GGMs here by 
leveraging the results of \cite{meinshausen2006high}. % and \cite{bondell2010joint}. 

{MNS} looks to accurately learn the set of highly variable edges
across a cohort of subjects whilst simultaneously
obtaining more reliable 
network estimates both at the population and subject-specific level.
This is achieved by proposing a more sophisticated model for the 
underlying covariance structure. The proposed model looks to
decompose covariance structure as the union of population 
effects which are reproducible across subjects with 
subject-specific idiosyncrasies. 
In this way, the {MNS} algorithm is able to leverage information across subjects
in a discriminative manner. This is in contrast to 
many of the current methodologies which leverage information 
in an indiscriminate fashion (e.g., via the use of group or fused regularization
penalties whose parameterization is fixed across edges).

Throughout a series of simulation studies we demonstrate that  
{MNS}  is able to accurately recover 
population and subject-specific functional connectivity networks.
The {MNS} algorithm is also shown to be capable to recovering 
the set of highly variable edges.
The simulations also provide empirical evidence 
of our claim that improved estimates of connectivity networks can 
be obtained by leveraging information across subjects in a discriminate manner. 
Moreover, a novel algorithm for simulation multiple related networks is proposed for the simulations.
While there is a wide number of algorithms available to simulate individual networks \citep{rubinov2010complex}, 
the question of simulating multiple related networks has received limited attention. 

The {MNS} algorithm requires the specification of two regularization
parameters, $\lambda_1$ and $\lambda_2$, each of which has a natural interpretation. 
The first parameter controls the sparsity in the 
 population node topologies while the second 
controls the sparsity of the subject-specific edges. 
We propose to learn both parameters via cross-validation.
Although the use of cross-validation incurs an additional 
computational cost, our experience is that 
it results in more accurate estimates of functional connectivity through improved 
selection of regularization parameters. 
However, a shortcoming of the proposed method is that 
both regularization parameters are dependent on each other \citep{schelldorfer2011estimation}.
Thus a large choice of $\lambda_1$ can be offset by a small choice of $\lambda_2$,
leading fixed effects to be explained by inflated random effects estimates.
% This poses concerns in the interpretation of the model output.
An additional shortcoming is that 
estimated functional connectivity networks must be inferred via the use of an 
AND or OR rule \citep{meinshausen2006high}. While such an approach 
is able to recover the sparse support structure, 
it is not able to obtain an estimate of the corresponding sparse precision matrix
which alternative methods such as the JGL are capable of obtaining.
Despite these shortcomings, we believe the additional richness of information 
provided by the {MNS} algorithm will make it invaluable in further understanding functional 
connectivity.

The MNS algorithm, together with network simulation methods described in this work have been implemented as an R package 
named \verb+MNS+.
This can be downloaded from the Comprehensive R Archive Network (CRAN). %\citep{CRANref}. 

{MNS}  was applied to resting-state fMRI data
taken from the ABIDE consortium \citep{di2014autism}.
The results serve to corroborate a widely held hypothesis within the 
neuroimaging community regarding the effect of subject movement of functional 
connectivity \citep{power2012spurious, satterthwaite2012impact} and are
consistent with the literature \citep{deichmann2003optimized}.
As far as we are aware, this is the first work to 
formally quantify variability in functional connectivity across  a cohort of subjects. 
In future, MNS could also be employed to study 
variability across populations including healthy controls and subjects suffering from
certain neuropathologies. Furthermore, another interesting 
avenue would be to employ MNS to study variability during task-based experiments. It has been
suggested that connectivity in this context is non-stationary \citep{hutchison2013dynamic, monti2014estimating, monti2015graph}.
Extending MNS to this scenario would allow for the study 
of variability induced by cognitive tasks.

In conclusion, the {MNS} algorithm provides a method for simultaneously
estimating population and subject-specific functional connectivity whilst also
reporting highly variable edges across a population.

\newpage
\appendix

% \section{Appendix section}\label{app}

\section{Stability approach for \textit{replicated} data}
\label{app--Stability}

In this section we briefly overview a stability selection (i.e., bootstrap) approach
for studying \textit{replicated} data. This approach is inspired 
by the $R^3$ approach proposed in \cite{narayan2015two}, however, it 
has a fundamentally different objective. As a result, some adjustments are 
introduced. 

As in the $R^3$ method, this approach is based upon resampling, randomized penalizations and 
random effects. The method, described in Algorithm \ref{algo_R3}, 
proceeds by iteratively obtaining bootstrapped estimates of covariance structure 
for each subject. These results are subsequently incorporated into a 
Beta-Binomial random effects model. Each of these steps is described below, for further
discussion and motivation of these steps we refer the reader to
\cite{narayan2013randomized} and \cite{narayan2015two}.

\subsection{Resampling}
In order to obtain reliable estimates 
of covariance structure the bootstrap is employed;
resulting in $B$ bootstrap estimates of connectivity structure per subject. Recall that 
the dataset for the $i$th subject, $X^{(i)} \in \mathbb{R}^{n \times p}$, consists of
$n$ observations across $p$ nodes.
At the $b$th bootstrap iteration, $n$ observations are sampled with replacement 
in order to form a bootstrapped dataset, $X^{(i), b} \in \mathbb{R}^{n \times p}$,
which is subsequently used to obtain an estimate for the covariance structure
using the Graphical Lasso, as described in Section \ref{app--RanPen}.  %\citep{friedman2008sparse}. 

\subsection{Randomization penalization}
\label{app--RanPen}

In order to alleviate possible bias introduced by the use of an $\ell_1$ penalty we employ 
randomized penalization techniques, an approach first introduced by \cite{meinshausen2010stability}.
The objective of randomized penalization schemes is to reduce the influence of
inclusion/exclusion of any edge on the presence of remaining edges.
Thus when estimating the network for the $b$th bootstrap sample, a random, symmetric penalty 
matrix, $\Lambda^{(i),b} \in \mathbb{R}^{p \times p}$, is employed. 

In order to obtain $\Lambda^{(i),b}$, we first estimate the regularization parameter for the $i$th subject
using the StARS method of 
\cite{liu2010stability}. This is performed only once using the entire (non-bootstrapped) dataset, $X^{(i)}$,
and is denoted by $\lambda^{(i)} \in \mathbb{R}$.
The randomized penalization matrix is  defined as follows:
\begin{equation}
\left (\Lambda^{(i),b} \right)_{k,j} =  \left (\Lambda^{(i),b} \right)_{j,k} = \lambda^{(i)} + c \lambda^{(i)}_{max} W_{j,k} \hspace{5mm} \forall j < k ,
\label{RanPenMatrix}
\end{equation}
where $\lambda_{max}^{(i)}$ is the value of sparsity parameter leading to a null model and $W \in \{-1,+1\}^{p \times p}$
is defined as:
\[
    W_{j,k}= 
\begin{cases}
    +1,& \text{w.p. }~ 0.5\\
    -1,              & \text{w.p. }~ 0.5
\end{cases}
~.\]

We are then able to obtain a penalized estimate of the precision as follows:
\begin{equation}
 \Theta^{(i),b} = \underset{\Theta }{\operatorname{argmin}} \left \{ - \mbox{log det}~\Theta + \mbox{trace} \left ( \frac{1}{n} {X^{(i),b}}^T X^{(i),b} \Theta \right ) + || \Lambda^{(i), b} \circ \Theta ||_1 \right \},
\label{RanPenTheta}
 \end{equation}
where $\circ$ denotes element-wise multiplication. 

\subsection{Random effects}

Finally, we look to formally quantify the presence or absence of edges at a population level. 
In order to achieve this a Beta-Binomial model is employed. 
For the $i$th subject we treat the presence of any given edge at each bootstrap iteration as a Binomial random variable.
We thus define $Y^{(i),B} \in \mathbb{R}^{p \times p}$ such that 
\begin{equation}
 Y^{(i),B}_{j,k} = \frac{1}{B} \sum_{b=1}^B \mathbb{I} \left (\Theta^{(i),b}_{j,k} \neq 0 \right ).
\end{equation}
Following \cite{narayan2015two}, 
$Y^{(i),B}_{j,k}$ is modeled as follows:
\begin{equation}
 Y^{(i),B}_{j,k} | \mu^{(i)}_{j,k} \sim \mbox{Binomial}~(\mu^{(i)}_{j,k}, B) \hspace{5mm} \mbox{and} \hspace{5mm} \mu^{(i)}_{j,k} \sim \mbox{Beta}~(\mu^{pop}_{j,k}, \rho^{pop}_{j,k}),
\end{equation}
where $\mu^{pop}_{j,k}$ is the population mean and $\rho^{pop}_{j,k}$ the variance. 
They can be  estimated as follows:
\begin{align}
\label{mupop}
 \mu^{pop}_{j,k} &= \frac{1}{N} \sum_{i=1}^N  Y^{(i),B}_{j,k} \\
 \rho^{pop}_{j,k} &= \frac{B}{B-1} ~\frac{ \sum_{i=1}^N \left (\mu^{pop}_{j,k} -Y^{(i),B}_{j,k} \right )^2}{\mu^{pop}_{j,k} (1-\mu^{pop}_{j,k})(N-1)} - \frac{1}{B-1}
 \label{rhopop}
 \end{align}

 These parameters are
subsequently used to infer population networks (using $\mu^{pop}$) as well as report highly
variable edges (using $\rho^{pop}$). Pseudo-code for the \textit{Stability} approach is provided in Algorithm \ref{algo_R3}.

 \begin{algorithm}[h!]
 \DontPrintSemicolon
 \KwIn{Data across $N$ subjects, $\{X^{(i)}\}$, number of bootstrap samples to perform, $B$.}
%  \KwResult{ $\mu^{pop}$, $\rho^{pop}$}
%  initialization\;
 \Begin{
 \For{i $\in \{1, \ldots, N\}$}{
 Select $\lambda^{(i)}$ using the StARS method \citep{liu2010stability}
 
 \For{$ b\in \{1, \ldots, B\}$}{
 Obtain $X^{(i),b}$ be sampling $n$ times with replacement from $X^{(i)}$
 
 Set randomization penalization matrix, $\Lambda^{(i),b}$, as in equation (\ref{RanPenMatrix})
 
 Estimate penalization precision matrix, $\Theta^{(i),b}$, as in equation (\ref{RanPenTheta})
 }
 }
 Estimate $\mu^{pop}$, $\rho^{pop}$ using equations (\ref{mupop}) and (\ref{rhopop})
 }
 \Return{$\mu^{pop}$, $\rho^{pop}$}
 \caption{Stability approach for \textit{replicated} data}
 \label{algo_R3}
\end{algorithm}

 \section{Simulating a cohort of networks}
\label{app--Simulating}
Producing synthetic data where the true underlying covariance structure is known is fundamental 
to providing an empirical validation of any algorithm. Moreover, in our scenario is it is crucial to ensure the 
simulated data demonstrates many of the properties often encountered in neuroimaging datasets. 
On a subject-specific level, this corresponds to a power law distribution of edges across nodes together with 
the presence of hub nodes. Such networks can be efficiently simulated using the scale-free network 
algorithm of \cite{barabasi1999emergence}. 

However, the problem of simulating multiple networks for multiple related subjects has not been 
thoroughly considered in the literature. While there is a vast literature on the properties which 
can be expected for subject-specific networks (see e.g., \cite{bullmore2009complex}), there is limited knowledge 
of the behavior which can be expected across a cohort of related subjects. 
In this work we look to address this issue by empirically studying resting state data
from healthy subjects taken from the ABIDE consortium \citep{di2014autism}.

The dataset employed here consisted of a resting state scan for $N=58$ subjects. For each subject, $n=230$ 
BOLD measurements were collected
over $p=92$ ROIs. We consequently estimated functional connectivity networks for each subject independently 
while employing a Graphical Lasso (Glasso) penalty. A stability selection procedure was employed, whereby a 
block bootstrap was used to resample the data for each subject multiple times. 
Randomized penalization was also employed to further correct for systematic bias in
model selection. 
% As a result, the proposed method shares many similarities to 
% the $R^3$ methods of \cite{narayan2015two}.
At each iteration, 
selected edges were recorded.
This allowed us to obtain a network per subject by selecting only 
edges that where consistently present across all iterations. A population network was also obtained 
by selecting the edges that where consistently present across all subjects. 
 
The properties of the resulting networks where studied in the hope 
of obtaining notable properties which could subsequently be recreated synthetically. 
In particular, we chose to focus on graph theoretic measures as these can be easily interpreted 
and calculated on simulated data \citep{rubinov2010complex}. Specifically, the clustering coefficient was studied across 
all networks. This provides a measure of how tightly nodes in a network tend to group together and 
expresses the cohesiveness of a network \citep{barrat2004architecture}. 
 
\begin{figure}[h!]
  \centering
\includegraphics[width=.8\textwidth]{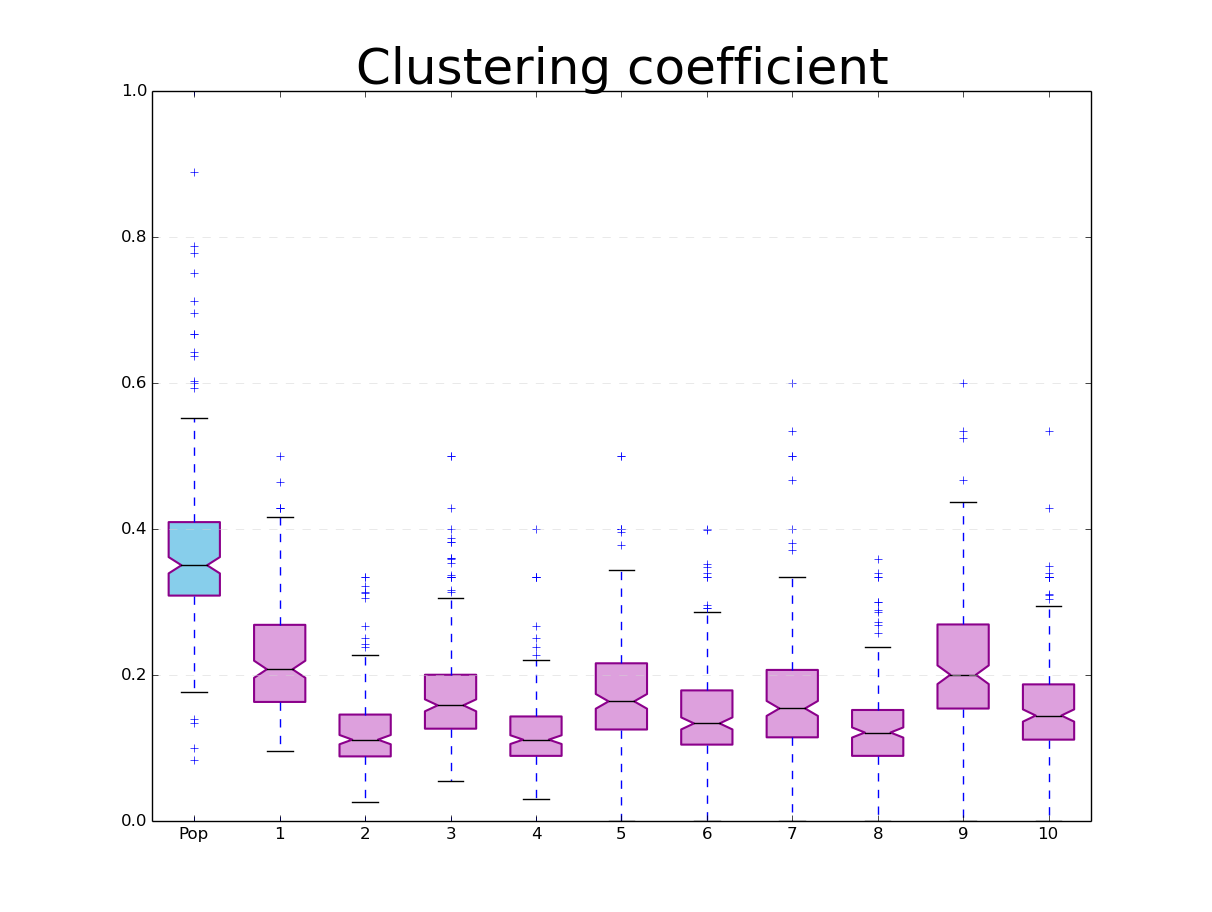}
\caption{Clustering coefficients for the population network (in blue) as well as 
for 10 randomly selected subjects. We note there is a clear drop in clustering coefficient from the
population network to the subject-specific networks.}
\label{ClusterCoef}
\end{figure}

 The results, shown in Figure [\ref{ClusterCoef}], show that the clustering coefficient is significantly
larger in the population network when compared the each of the subject-specific networks. 
We hypothesize that this is a manifestation of the fact that there is a highly structured
population network present. We further hypothesize that it is the large 
inter-subject variability which accounts for some of the drop in clustering coefficient at 
the subject-specific level. 

When simulating synthetic networks, as described in Algorithm \ref{algo1}, we look to 
recreate these properties (i.e., the drop in clustering coefficient). This is achieved 
by first simulating a population network according to the \cite{barabasi1999emergence} model. 
This results in a highly structure population network which also demonstrates many of the properties
known to be present in neuroimaging data (e.g., power law distribution and the presence of hub nodes).
A subset of highly variable edges, denoted by $\tilde E$, is then randomly selected according to
the \cite{erdHos1959random} model. For each subject,
each edge in $\tilde E$ is added to the subject-specific network with a given probability, $\tau$.
This yields variable edges that are only 
present across a subset of the population. The introduction of these random edges serves to reduce the clustering 
coefficient of the subject-specific network, thereby recreating the 
properties observed in our exploratory analysis.

\subsection{Ensuring positive definiteness}
 
Through algorithm \ref{algo1} we are able to simulate 
a population precision, $\Theta^{pop}$,
together with subject-specific deviations, $\Theta^{(i)}$.
We define the precision for each subject to
be $\Theta^{pop} + \Theta^{(i)}$, however, care must be taken to 
ensure this sum is positive-definite. 
In this work we follow \cite{danaher2014joint} and 
ensure the subject-specific precision matrices are
positive definite by rescaling the matrix.
Formally, each off-diagonal element is divided by
the sum of the absolute values of all off-diagonal elements in its row.
This yields a non-symmetric matrix which is subsequently averaged with 
its transpose.

\section{Further simulations}
\label{DanaherSims}

In Section \ref{sec--SimStudy} networks were simulated as described in Algorithm \ref{algo1}.
While this algorithm was derived from an exploratory analysis of resting-state fMRI data, a wide range of alternative 
algorithms could also be proposed. In this section we look to provide further empirical evidence 
by recreating the simulation study of \cite{danaher2014joint}. 

While \cite{danaher2014joint} are able to simulate networks where variability is present, their proposed method 
is designed primarily to provide empirical evidence on how accurately subject-specific networks could be recovered. 
We therefore follow \cite{danaher2014joint} and focus exclusively on recovering subject-specific covariance structure here. 

Two simulations where performed where data was simulated for $N=3$ subjects and 
$p=100$ and $p=250$ nodes respectively.
Within each simulation, nodes where divided into 10 equally sized and unconnected components. 
The connectivity structure within each component was simulated according to scale-free model of
\cite{barabasi1999emergence}, resulting in 10 scale-free sub-networks.
Of the 10 sub-networks, eight were shared across all three subjects.
Of the remaining two sub-networks, 
one was 
present in two out of three subjects and the final sub-network was only present in the first subject.
For further see \cite{danaher2014joint}.
 
\subsection{Simulation 1: $p=100$, $n \in \{50,100,200\}$}

\begin{figure}[h!]
  \centering
\includegraphics[width=\textwidth]{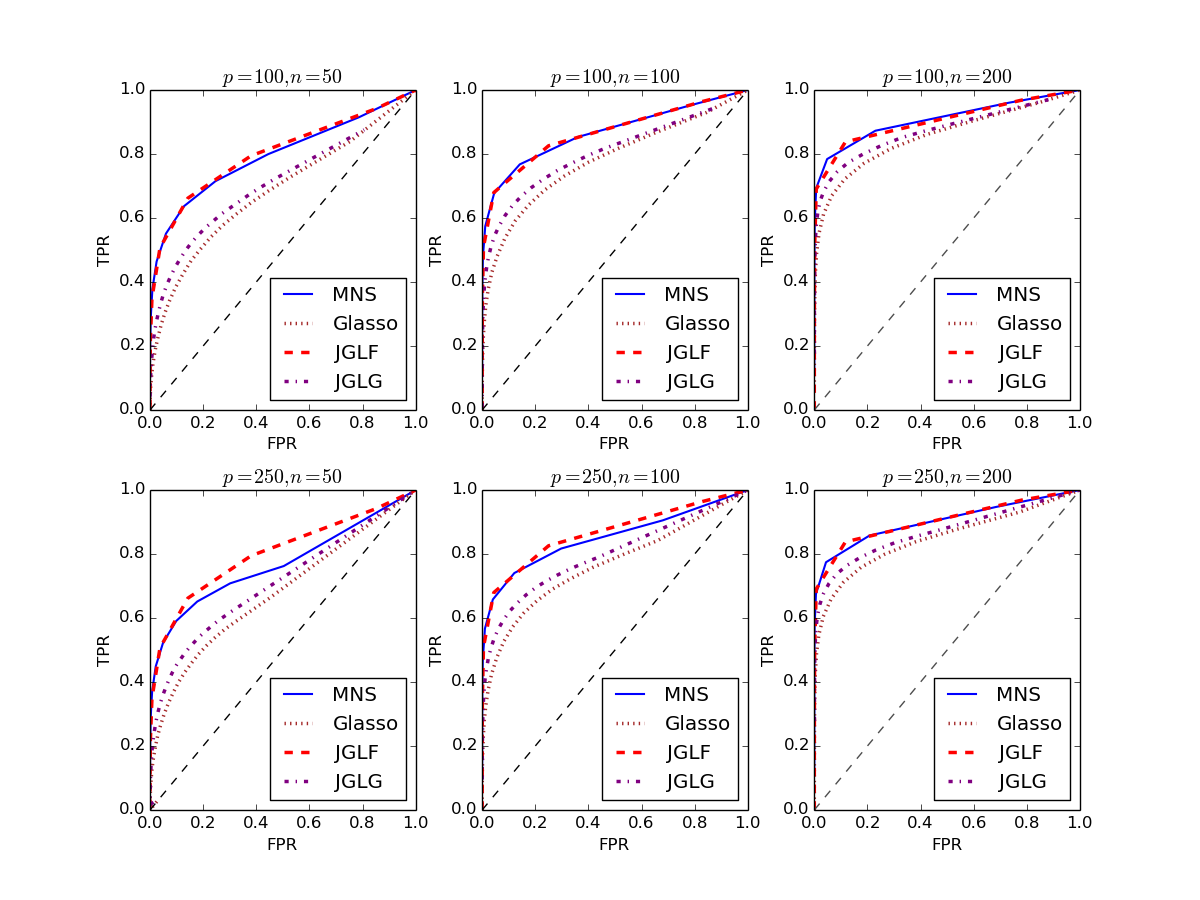}
\caption{Results for simulations 1 and 2. These simulations sought to re-create the simulation study 
presented in \cite{danaher2014joint} with $p=100$ and $p=250$ nodes respectively. }
\label{JGL1}
\end{figure}

In the first simulation $p=100$ nodes were employed resulting in 10 sub-networks each with 10 nodes.
The number of observations per subject was allowed to vary from $n=50$ through to $n=200$. 
The results over 100 simulations are shown in top row of  Figure [\ref{JGL1}]. The {MNS} algorithm performs 
competitively with respect to the JGL-Fused algorithm and 
outperforms both the JGL-Group and graphical lasso algorithms across all
values of $n$. In particular, we note that the {MNS} algorithm remains competitive even as the number of observations, $n$,
falls drastically.

\subsection{Simulation 2: $p=250$, $n \in \{50, 100, 200\}$}
 
 The second simulation employed $p=250$ nodes which were divided into 10 sub-networks 
of 25 nodes each.
The number of observations per subject was allowed to vary from $n=50$ through to $n=200$ as before. 
The results over 100 simulations are shown in bottom row of  Figure [\ref{JGL1}]. As before, the {MNS} algorithm
performs competitively against alternative algorithms. As with the previous simulation, we note there is a 
trend for ROC curves to improve as the number of observations, $n$, increases.

\begin{table}[ht]
\centering
\begin{tabular}{c|c|cc|cc|cc|cc}
  \hline
   $p$ & $n$ & \multicolumn{2}{c}{MNS} & \multicolumn{2}{c}{Glasso} & \multicolumn{2}{c}{JGL Fused} & \multicolumn{2}{c}{JGL Group}\\
   \hline 
  \multirow{3}{*}{100} 
   &50 & 0.346 & 0.006 & 0.175 & 0.016 & 0.343 & 0.007 & 0.221 & 0.012 \\ 
   & 100& 0.477 & 0.003 & 0.282 & 0.008 & 0.503 & 0.005 & 0.353 & 0.005 \\ 
   & 200 & 0.594 & 0.002 & 0.429 & 0.004 & 0.632 & 0.005 & 0.514 & 0.003 \\ 
%   & & & & & & &\\ 
\hline
  \multirow{3}{*}{250} &
  50 &  0.287 & 0.002 & 0.125 & 0.007 & 0.292 & 0.003 & 0.161 & 0.005 \\ 
   & 100 & 0.443 & 0.002 & 0.215 & 0.003 & 0.451 & 0.002 & 0.295 & 0.003 \\ 
   & 200 & 0.573 & 0.001 & 0.370 & 0.002 & 0.584 & 0.002 & 0.465 & 0.002 \\
   \hline
  
   \hline
\end{tabular}
\caption{Performance of all four algorithms when recovering subject
specific functional connectivity structure}
\end{table}

\section*{Acknowledgements}
The authors wish to acknowledge the help of Dr. Gareth Ball in preprocessing the data.

% \begin{supplement}
% \sname{Supplement A}\label{suppA}
% \stitle{Title of the Supplement A}
% \slink[url]{http://www.e-publications.org/ims/support/dowload/imsart-ims.zip}
% \sdescription{Dum esset rex in
% accubitu suo, nardus mea dedit odorem suavitatis. Quoniam confortavit
% seras portarum tuarum, benedixit filiis tuis in te. Qui posuit fines tuos}
% \end{supplement}

\newpage 
\bibliographystyle{plainnat}
\bibliography{ref}

\end{document}